\documentclass{article}

\usepackage[preprint,nonatbib]{neurips_2024}

\usepackage{amsmath}
\usepackage{amssymb}
\usepackage{mathtools}
\usepackage{amsthm}
\usepackage{subfig}

\usepackage[utf8]{inputenc} 
\usepackage[T1]{fontenc}    

\usepackage{url}            
\usepackage{booktabs}       
\usepackage{amsfonts}       
\usepackage{nicefrac}       
\usepackage{microtype}      
\usepackage{xcolor}         
\usepackage{multirow}

\theoremstyle{plain}
\newtheorem{theorem}{Theorem}
\newtheorem{proposition}{Proposition}
\newtheorem{lemma}{Lemma}

\theoremstyle{definition}

\newtheorem{property}{Property}
\newtheorem{assumption}{Assumption}
\newtheorem{condition}{Condition}
\theoremstyle{remark}

\usepackage[style=numeric,maxnames=1,minnames=1]{biblatex}
\usepackage[colorlinks=true,linkcolor=blue,citecolor=blue,urlcolor=blue]{hyperref} 

\usepackage{graphicx} 
\usepackage{wrapfig}
\usepackage{cleveref}

\newcommand{\myref}[2]{%
  \hyperref[#1]{#2}
}  

\usepackage{algorithm}  
\usepackage{algorithmic}

\title{IENE: Identifying and Extrapolating the Node Environment for Out-of-Distribution Generalization on Graphs}

\author{%
  Haoran Yang, Xiaobing Pei, Kai Yuan
    \\
  Department of Software Engineering\\
  Huazhong University of Science and Technology\\
  \texttt{\{m202276630, xiaobingp, m202376878\}@hust.edu.cn} \\
}

\begin{document}

\maketitle

\begin{abstract}
Due to the performance degradation of graph neural networks (GNNs) under distribution shifts, the work on out-of-distribution (OOD) generalization on graphs has received widespread attention. A novel perspective involves distinguishing potential confounding biases from different environments through environmental identification, enabling the model to escape environmentally-sensitive correlations and maintain stable performance under distribution shifts. However, in graph data, confounding factors not only affect the generation process of node features but also influence the complex interaction between nodes. We observe that neglecting either aspect of them will lead to a decrease in performance. In this paper, we propose IENE, an OOD generalization method on graphs based on node-level environmental identification and extrapolation techniques. It strengthens the model's ability to extract invariance from two granularities simultaneously, leading to improved generalization. Specifically, to identify invariance in features, we utilize the disentangled information bottleneck framework to achieve mutual promotion between node-level environmental estimation and invariant feature learning. Furthermore, we extrapolate topological environments through graph augmentation techniques to identify structural invariance. We implement the conceptual method with specific algorithms and provide theoretical analysis and proofs for our approach. Extensive experimental evaluations on two synthetic and four real-world OOD datasets validate the superiority of IENE, which outperforms existing techniques and provides a flexible framework for enhancing the generalization of GNNs.
\end{abstract}

\section{Introduction}
\label{introduction}

Graph Neural Networks (GNNs) are highly effective deep learning algorithms specifically designed for processing graph-structured data, exhibiting various forms of variations \cite{Hamilton,Kipf,Velickovic}. One primary task of GNNs involves performing node-level predictions on graphs, which has numerous applications in recommendation systems \cite{Wu01,zhang}, fraud detection \cite{Li01}, and social networks \cite{Wang,Yin}.

The success of the learning paradigm of GNNs relies on the assumption that the data follows the principle of independent and identically distributed (IID). Nevertheless, in practical scenarios, it is often challenging to satisfy this assumption due to various uncontrollable factors inherent in real-world data generation mechanisms, such as data selection biases, confounding factors, and other characteristics\cite{Bengio,Engstrom,Su}. The test distribution may introduce uncontrollable deviations, commonly referred to as out-of-distribution (OOD) shifts \cite{Jin,Krueger}, which frequently results in unstable predictions for most GNNs. In essence, the primary cause of accuracy degradation in GNNs is the existence of accidental correlations between spurious features and class labels. Models trained using Empirical Risk Minimization (ERM) often struggle to distinguish spurious features and may instead rely on environmental cues for classification. Consequently, any environmental changes may significantly degrade the performance of the classifier. \cite{Li02}. 

To tackle the OOD problem, a widely-adopted approach aims to extract invariant features \(X_{i}\) that exhibit a stable correlation with the target \textit{Y} and can reliably predict \textit{Y} in new test distributions \cite{Arjovsky,Kuang,Rosenfeld,Shen}. The recent advancements in invariant learning have also followed this principle \cite{Lin01,Tan,Zhou}. In general, invariant learning assumes the existence of multiple discrete environments in the training dataset. In different environments, the conditional distribution \(P(Y|X_{s})\) differs, but \(P(Y|X_{i})\) remains invariant. From a causal perspective, \(X_{i}\) is the direct cause of \textit{Y}, while other factors are represented by \(X_{s}\) (further discussion in Section \ref{section3}). Invariant learning techniques aim to train a feature extractor \(\Phi\) that focuses solely on \(X_{i}\) without being influenced by \(X_{s}\). This implies that in any environment, a representation such as \(\Phi \left ( X \right )\) will yield equal (optimal) performance for downstream classifiers. Existing research has demonstrated that \(\Phi\) can extract \(X_{i}\) across a sufficient number of distinct environments and under other appropriate conditions. Nevertheless, in practical scenarios, it is often difficult to satisfy the assumption of environmental heterogeneity without prior knowledge of the partitioning of environments, as such information is often unavailable \cite{Creager,Liu01}. 

To achieve environmental heterogeneity in features, a crucial approach is environment partitioning, which  involves constructing data for multiple environments by dividing the node sets into independent subsets. Some prior studies have attempted to achieve environmental recognition by making additional assumptions about \(X_{s}\). For example, \textcite{Creager} assume that the ERM method learns only \(X_{s}\), while \textcite{Liu01} assume that the differences in cluster features are greater than those in invariant features. However, these assumptions are difficult to validate as they may not universally hold, and \(X_{s}\) may not be the direct cause of \textit{Y}. Although ZIN provides an environment prediction method based on \(X_{z}\), avoiding additional assumptions about spurious features, it explicitly requires prior knowledge (i.e., \(X_{z} \bot Y|X_{i}\)) and working conditions (i.e., given latent \(X_{z}\)). However, obtaining such prior knowledge can be challenging in practical scenarios, thus limiting the broader application of the method. Therefore, TIVA \cite{Tan}, attempted to automatically learn environmental partitions using target-independent variables \(X_{ir}\), but the underutilization of information may lead to the suboptimal performance. BA-GNN \cite{Chen01} identifies environments by clustering nodes based purely on the relationships between their variant information and target labels, which can not guarantee that invariant information is not separated. CaNet \cite{QTWu} utilizes parameterized techniques to predict pseudo-environments at each layer of the GNN, which resemble the gating scores in Mixture of Experts (MoE) and differ from the environment we defined.

In addition to node features, structural information can also impact the model's ability to generalize under environmental shifts, which introduces unique technical challenges in addressing distribution shifts on graphs. To discover invariance in structural shifts across environments, recent efforts involve generating augmented views and conducting invariant learning in multi-environment settings. For instance, FLOOD \cite{Liu02} and Lisa \cite{Lisa} employs general graph augmentation techniques such as random node feature masking \cite{You} and random DropEdge \cite{Rong}. CAT \cite{He} generates structural interventions by minimizing structure correlations related to node clustering embeddings. EERM \cite{Wu02} trains multiple graph structure generators to maximize risk divergence from various virtual environments. However, there is no guarantee that these generated views adequately satisfy the assumption of environmental heterogeneity, as merely introducing varying degrees of noise or reducing information to different extents can also increase risk divergence in different environments.

In this work, we first propose an automatic approach for learning environmental partitioning in a data-driven manner. To achieve this, we introduce a feature disentanglement framework to simultaneously learn environmental representation \(h_{e}\) and invariant representation \(h_{i}\). We utilize \(h_{e}\) as auxiliary information for environment partitioning and employ \(h_{i}\) for invariant learning. Furthermore, to promote the learning of invariance in structure, we develop a novel environmental extrapolation method to generate multiple augmented views and utilize NV-REx, a node-level invariance penalty, to minimize the risk discrepancy across different environments. Our framework integrates invariant learning strategies at both feature and structural granularities, enabling the discovery of invariant patterns when environmental changes occur on graphs. In Section \ref{section3.4}, we theoretically demonstrate the ability of IENE to identify invariance. Subsequently, we conducted practical experiments on both synthetic and real-world datasets to validate the superiority of IENE. 
We summarize our contributions as follows:
\begin{itemize}
\item We propose IENE, an approach that integrates environmental identification, environmental extrapolation techniques, and an invariance learning framework to discover invariant patterns on graphs, enabling generalization of GNNs when distribution shifts occur on the graph.

\item we apply disentangled information bottleneck framework to the alternating learning of environmental inferring and invariance in features, and discover invariance in structures through environmental extrapolation to identify more accurate invariant and variant patterns.  This is a novel insight in the current research on invariant learning on graphs.

\item We theoretically prove that IENE can identify invariant features on graphs under appropriate conditions. And our empirical evaluation demonstrates that, compared to baseline methods, IENE exhibits superior performance on multiple synthetic and real-world datasets.
\end{itemize}
Organization of the rest of the paper. We begin by presenting preliminary work in Section \ref{section2}. Subsequently, we introduced the two main modules of IENE, conduct theoretical analysis, and describe the specific algorithm in Section \ref{section3}. Then, in Section \ref{section4}, we evaluate the proposed method by comparing it with several baseline methods on both synthetic and real OOD datasets. Finally, we summarize the work and discuss future directions in Section \ref{section5}. In \myref{Appendix}{Appendix}, we provide alternative algorithms for IENE, detailed theoretical analyses and proofs, additional ablation studies, experimental details and reviews of related work.

\section{Preliminaries}
\label{section2}
\vspace{-7pt}
\subsection{Node-Level Environment Extrapolation Problem for OOD Generalization}

Assuming \(v\in V\) is a random variable representing a node. Similar to EERM \cite{Wu02}, we use an ego-graph to define the subgraph centered on a node. We define the  \textit{L}-th order inner neighbor set of node \(v\) as \(N_v\) (including \(v\) itself), where the nodes and connections in \(N_v\) constitute the ego-graph \(G_v\) of \(v\). The ego-graph is comprised of a local node feature matrix \(X_v=\left \{x_u|u\in N_v  \right \} \) and a local adjacency matrix \(A_v=\left \{a_{uw}|u,w\in N_v  \right \} \). Using \(\boldsymbol{G_v}\) as the random variable for the ego-graph, its implementation is denoted as \(G_v=\left ( A_v,X_v \right ) \). Additionally, we can represent the entire graph as a set of local graph instances \(G=\left \{ G_v \right \},v\in  V\). The ego-graph can be seen as the Markov blanket of the central node, and the conditional distribution \(p\left ( \boldsymbol{Y}|\boldsymbol{G},\boldsymbol{e} \right )\) can be decomposed as a product of \(|V|\) independent and identical marginal distributions \(p\left ( \boldsymbol{y}|\boldsymbol{G_v},\boldsymbol{e_v} \right )\). 

We define \(\boldsymbol{e}\) as a random variable for the node environment, where \(e\in \mathbb{R}^{|V|\times K} \) represents a specific instance of \(\boldsymbol{e}\). The node-level OOD problem can be formulated as follows: 

\textbf{Problem 1}: Given training data from distribution \(p\left ( \boldsymbol{Y},\boldsymbol{G}|\boldsymbol{e}=e \right )\), the model needs to handle test data  from a new distribution \(p\left ( \boldsymbol{Y},\boldsymbol{G}|\boldsymbol{e}=e' \right )\). Using \(\varepsilon=supp(\boldsymbol{e}) \) to denote the support set of the environment, \(f\) for the prediction model, and \(l\left ( \cdot |\cdot  \right ) \) for the loss function, more formally, the OOD problem can be written as:
{
\abovedisplayskip=12pt 
\belowdisplayskip=12pt 
\begin{equation}
\label{equation1}
\begin{aligned}
&f^{*} =\min_{f} \max_{e\in \varepsilon} \mathbb{E}_v[\mathbb{E}_{e_v}[l(f(G_v),y_v)]]
\end{aligned}
\end{equation}
}
Eqn. (\ref{equation1}) is inherently unsolvable in the natural scenario, with no prior knowledge or structural assumptions to obtain \textit{e} that maximizes the inner term, as access is limited to environments in the training set. However, assuming that environments are identifiable and assumption \ref{assumption5} holds, we can extend the training environment to diverse environments by extrapolation.

In summary, the node-level OOD problem can be approximately addressed by extrapolating node environments based on graph augmentation, and conducting invariant learning on graphs under multiple environments for OOD generalization. Furthermore, \textbf{Problem \ref{equation1}} extends to \textbf{Problem \ref{equation2}}.

\textbf{Problem 2}: Given a graph dataset \(G= \left\{ X,A|e \right\}\), with \(G_v^{e_k} \) representing the ego-graph centered at node \(v\) under environment \(e_k\). The task is to generate distributions \(P\left(G|\boldsymbol{e}=\overline{e}_k \right)\) from a given distribution \(P\left(G|\boldsymbol{e}=e\right)\). In other words, generate multiple new graphs \(\{G_k\}=\{m\left(G,n_k\right)\}\), where \(m(\cdot|\cdot)\) is a non-linear mapping function satisfying assumption \ref{assumption5}, and \(n_k\) is a learnable independent noise variable for environment \(\overline{e}_k\). $\{G_k\}$ augment the observed environments during training, leading to $\varepsilon_{tr}\to \varepsilon_{ex}$. Assuming \(\overline{e}_k \in \varepsilon_{ex}\) and \(\overline{e}_j \in \varepsilon_{ex}\setminus \overline{e}_k\), if \({e}_k\perp {e}_j\), then node environment extrapolation achieves data distribution augmentation for \(\varepsilon_{tr}\). Subsequently, learn a predictor \(f^*\) defined in problem \ref{equation1} to achieve better OOD generalization performance.
{
\abovedisplayskip=12pt 
\belowdisplayskip=12pt
\begin{equation} 
\label{equation2}
\begin{aligned}
&f^{*} =\min_{f} \max_{e\in \varepsilon_{ex}} \mathbb{E}_v[\mathbb{E}_{e_k}[l(f(G_v^{e_k}),y_v)]]\\&
s.t.\ G_v^{e_k}=m(G_v,n_k)
\end{aligned}
\end{equation}
}

\subsection{Environment Identification Problem}
Assuming that node features \(X_v\in\mathbb{R}^{|N_v|\times d}\) in ego-graph \(G_v\) are composed of invariant features \(X_{v_{i}}\in\mathbb{R}^{|N_v|\times d_{i}}\), spurious features \(X_{v_{s}}\in\mathbb{R}^{|N_v|\times d_{s}}\), and irrelevant features \(X_{v_{ir}}\in\mathbb{R}^{|N_v|\times d_{ir}}\) by function \(q\), i.e. \(X_v=q(X_{v_{i}},X_{v_{s}},X_{v_{ir}})\). Moreover, the target \(y\) is generated by a non-degenerate function \(g_{i}\) with independent random noise: \(y=g_{i}(X_{v_{i}},A_v,\epsilon _v)\), where \(X_{v_{i}}\perp \epsilon _v \). The probability distribution \(P(y|X_{v_{s}})\) changes in different environments, while \(P(y|X_{v_{i}})\) remains invariant.

According to the analysis in \ref{B1},\(X_{v_{i}}\) is not beneficial to environment identification and may even introduce additional noise, while \(X_{v_{s}}\) and \(X_{v_{ir}}\) play crucial roles in environmental inference. Assuming the environment recognition module \(\varphi\) is composed of a environmental feature extractor \(u\) and an environment classifier \(w\), i.e., \(\varphi(\cdot)=w(u(\cdot))\). During environment identification, following \textcite{Lin01}, we consider that graph data is not collected from a single discrete environment but from a mixture of multiple environments, with the distribution \(P(X,A,Y)= {\textstyle \sum_{e\in\varepsilon }^{}}\alpha^eP(X^e,A^e,Y^e) \), where \(\alpha^e\in [0,1]\) and \( {\textstyle \sum_{e\in \varepsilon }^{}} \alpha ^e=1\). The goal of environment identification is to learn a feature representation \(h_{e}\) to predict \(\alpha\). 

\section{Method}
\label{section3}
\vspace{-7pt}
\subsection{Identifying Invariance in features through Environment Partitioning}
\label{section3.2}
As \textcite{Lin01} demonstrated that it is impossible to perform generally environment partitioning without introducing predefined environmental knowledge or auxiliary information, our first step is to extract auxiliary information from the input graph for environment partitioning. Here, we consider mapping the ego-graph \(G_v\) to the environmental representation \(h_{e}\) through a learning function \(u(\cdot):\mathbb{R}^{|N_v|\times d}\to \mathbb{R}^H\), where \(H\) is the dimension of the hidden layer. 

A main challenge is how we learn \(h_{e}\). From a causal perspective, \(h_{e}\) and \(h_{i}\) belong to two independent subspaces (\(h_{e}\perp h_{i}\)) of \(G_v\) \cite{Chen03,Von}, and \(h_{e}\) should contain information relevant to environmental prediction. Here, we leverage information beyond the invariant features in the raw data as a substitute for \(h_{e}\). Adjusting the objective function for Disentangled Information Bottleneck \cite{Pan}, we learn \(h_{e}\) by minimizing the following equation: 
{
\abovedisplayskip=6pt 
\belowdisplayskip=6pt 
\begin{equation}
\begin{aligned}
\label{equation3}
&L_{DisenIB}=-I(h_{i};Y)
-I(G_v;h_{e},h_{i})+I(h_{e};h_{i})\\
\end{aligned}
\end{equation}
}
For the specific implementation algorithm of Eqn. (\ref{equation3}), refer to Section \ref{section3.5.2}. Note that \(h_{v_{e}}\) may contain noise unrelated to environmental prediction, and we will discuss how to leverage \(h_{v_{e}}\) for effective environment partitioning in the following. 

The goal of environment recognition is to learn a discriminator that predicts environment partition based on \(h_{v_{e}}\). Considering \(K\) as a tunable hyperparameter, we assume another function \(w(\cdot):\mathbb{R}^{H}\to \mathbb{R}^K\) can map the environmental representation \(h_{v_{e}}\) to a \(K\)-dimensional vector \(\rho \). Inspired by \textcite{Lin01}, data is generated by a mixture of multiple environmental factors. Based on this perspective, we have \(\rho^{(k)} \in [0,1] \), and \(\sum \rho^{(k)}=1\), where \( \rho^{(k)} \) is the \(k\)-th entry of \(\rho \).

The goal of IRM \cite{Arjovsky} is to learn an invariant feature extractor \(\Phi\), which extracts the same representation across all environments and provides equal (optimal) performance for downstream classifiers. To achieve this objective, we fit classifiers \(c_k\) in discrete environments  and a shared classifier \(c\) for multiple domains. Given \(\rho_{u,w}\), the invariant penalty is defined as:
{
\abovedisplayskip=6pt 
\belowdisplayskip=6pt 
\begin{equation}
\begin{aligned}
\label{equation4}
\mathcal{P}_{inv}(\rho_{u,w},\Phi,c,c_1,...,c_k)=
 {\textstyle \sum_{k=1}^{K}}\left [  
 \mathcal{R}^k_{\rho_{u,w}} (c,\Phi)
 -\mathcal{R}^k_{\rho_{u,w}} (c_k,\Phi)
 \right ]
\end{aligned}
\end{equation}
}
where \(\mathcal{R}^k_{\rho_{u,w}} (c,\Phi)=\frac{1}{n}  {\textstyle \sum_{v\in V}^{}} \rho ^{(k)}(G_v){l}(c(\Phi (G_v)),y_v)\). Inspired by \textcite{Tan}, we adopt the following min-max procedure to adversarially learn environment partitioning and extract invariant features. The objective of feature invariant learning is:
{
\abovedisplayskip=6pt 
\belowdisplayskip=6pt 
\begin{equation}
\begin{aligned}
\label{equation5}
\min_{c,\Phi } \max_{u,w,c_1,...,c_K}L_s:= 
\mathcal{R}(c,\Phi)
+\lambda \cdot \mathcal{P}_{inv}(\rho _{u,w},\Phi,c,c_1,...,c_K)
\end{aligned}
\end{equation}
}
where \(\mathcal{R} (c,\Phi)=\frac{1}{n}  {\textstyle \sum_{v\in V}^{}} l(c(\Phi (G_v)),y_v)\) represents the ERM loss. The inner max part in Eqn. (\ref{equation5}) attempts to find an environment partition where spurious features would incur the maximum penalty. 

\vspace{-4pt}
\subsection{Identifying Invariance in structures through Environment Extrapolation}
\label{section3.3}
\vspace{-3pt}
In Section \ref{section3.2}, we identified invariance through environmental partitioning by finding the partition that maximizes the penalty. However, according to analysis in \ref{B2}, the invariant features identified in static graph may struggle to adapt to structural changes in the environment. In this section, our goal is to identify dynamic invariant features that not only preserve the properties of invariance but also should be insensitive to structural changes. Based on environmental partitioning, we further leverage graph augmentation to achieve environment extrapolation and identify dynamic invariant features with an invariance objective.

Similar to the setting of \cite{Lin01}, we have \(\rho^{(k)} \in [0,1] \), and \(\sum \rho^{(k)}=1\), meaning that data is collected from a mixture of multiple environments with the distribution \(P(X,A,Y)= {\textstyle \sum_{k}^{}}\rho^{(k)}P(X^k,A^k,Y^k) \), where \(\rho^{(k)}\) represents the weight or probability of being collected from a specific environment. Environmental extrapolation implies reducing the coupling between spurious features from different environments, bringing the augmented graph closer to a discrete environment. Based on the above intuition, our goal for environmental extrapolation is:
{
\abovedisplayskip=6pt 
\belowdisplayskip=6pt 
\begin{equation}
\begin{aligned}
\label{equation6}
G_v^{(k)}= \mathop{\text{argmin}}\limits_{G_v^{(k)}}
\mathbb{E}_v  \left[   \mathbb{E}_k 
 \left[l(w(u({G}_v^{(k)} )),e^{(k)})  \right ] \right]
\end{aligned}
\end{equation}
}
where \(G_v^{(k)}=(X_v^{(k)},A_v^{(k)})\) represents the ego-graph of a node \(v\) under the environment index \(k\), and \(e^{(k)}\) represents a one-hot vector with only the \(k\)-th position being 1. Here, \(w\) and \(u\) are given after training in an adversarial manner according to Eqn. (\ref{equation5}), capable of automatically inferring the environment \(\varphi(G_v)=w(u(G_v))\) based on the input graph. We assume that \(w\) and \(u\) can approximately fit the generation process of data in each discrete environment with non-linear rules. By updating \(G_v^{(k)}\) by Eqn. (\ref{equation6}), we can approximate a series of new graphs \(G^{(k)}=\{G_v^{(k)}\},v \in V\) with maximal environmental differences, thus achieving environmental extrapolation.

For discrete multi-environmental data, some literature aiming at node-level OOD generalization \cite{Liu02,Wu02} directly adopts V-REx \cite{Krueger} as the invariance penalty. Based on analysis \ref{B3}, in some cases, V-REx may struggle to impose a sufficiently large penalty on differences, latently leading to suboptimal results. Therefore, we improved V-Rex, proposing a new invariance penalty called N(Node-wise)V-REx, as shown in Eqn. (\ref{equation7}), where \(f(\cdot)=c(\Phi(\cdot))\). In Appendix \ref{C}, it is proven that, compared to V-REx, it has a tighter upper bound for optimizing OOD errors measured by \(D_{KL}(p_{e'}(y|G_v)||q(y|G_v))\).
{
\abovedisplayskip=6pt 
\belowdisplayskip=6pt 
\begin{equation}
\begin{aligned}
\label{equation7}
\mathbb{V}(\mathcal L(G^{(k)},Y))=
\mathbb{E}_v \mathbb{E}_k  \left[ 
l (f(G_v^{(k)}),y_v) 
-\mathbb{E}_k \left[l(f(G_v^{(k)}),y_v) \right]
\right]^2
\end{aligned}
\end{equation}
}
where \(\mathcal L(G^{(k)},Y)=\mathbb{E}_v  
\mathbb{E}_k
l (f(G_v^{(k)}),y_v)\), and the objective of dynamic invariant learning is given by Eqn. (\ref{equation8}) for given \(G_v^{(1)},...,G_v^{(k)}\).
{
\abovedisplayskip=6pt 
\belowdisplayskip=6pt 
\begin{equation}
\begin{aligned}
\label{equation8}
&\min_{c,\Phi}L_d:=\mathcal{L}(\hat{G}^{(k)},Y)
+\beta \cdot \mathbb{V}(\mathcal{L}(\hat{G}^{(k)},Y) ) 
\end{aligned}
\end{equation}
}
In Section \ref{section3.4}, we discussed the sufficient conditions for identifying invariant features by Eqn. (\ref{equation5}) and (\ref{equation8}) without prior knowledge and using inductive biases. Subsequently, from a theoretical perspective, we proposed practical algorithms in Section \ref{section3.5}.

\subsection{Theoretical Analysis}
\label{section3.4}
\subsubsection{Why IENE can identify invariant features}
\label{section3.4.2}
First, we will demonstrate that the objective of Eqn. (\ref{equation5}) can ensure the identification of invariant features. For this, we rely on Assumptions \ref{assumption2}-\ref{assumption4} and Conditions \ref{condition1}-\ref{condition2}. Theorem \ref{theorem1} generalizes Theorem 2 of ZIN \cite{Lin01} to graphs.
\begin{theorem}
\label{theorem1}
Based on Assumptions \ref{assumption2}-\ref{assumption4} and Conditions \ref{condition1}-\ref{condition2}, if there exists \(\epsilon <\frac{C\gamma \delta }{4\gamma +2C\delta H(Y)}\) and \(\lambda \in [\frac{H(Y)+1/2\delta C }{\delta C-4\epsilon }-\frac{1}{2},\frac{\gamma }{4\epsilon }-\frac{1}{2}\), let \(\Phi^*\) be the solution to Eqn. (\ref{equation5}), then we have \(\hat{\mathcal{L}}(\Phi_i)>\hat{\mathcal{L}}(\Phi^*)\) for all \(\Phi_i \neq \Phi^*\), where \(H(Y)\) represents the entropy of \(Y\), and \(\hat{\mathcal{L}}(\Phi_i)\) represents the Loss when \(\Phi(\cdot)\) is \(\Phi_i\). Thus, the solution to Eqn. (\ref{equation5}) determines the invariant features. See Appendix \ref{C} for the proof.
\end{theorem}

The necessary conditions for identifiability have been sufficiently discussed in Appendix B.3-5 of ZIN \cite{Lin01}. It is worth noting that, since there exists a "invariant" spurious feature \(X_{s}^s\) in all possible environment partitions, adding this feature to \(\Phi(G)\) does not incur any invariance penalty. This indicates that through Eqn. (\ref{equation5}), what we learn are static invariant features \(X_{i}^s\), which actually encompass a subset of spurious features \(X_{i}^s=(X_{i},X_{s}^s)\). Environments should have sufficient diversity and informativeness so that each spurious feature can be identified and penalized. Therefore, we extend the environment partition to environment extrapolation, allowing for the identification of more invariant features from \(X_{i}^s\) that can stably predict \(Y\) under structural dynamic changes. 

Following the derivation route of EERM, we propose Theorem \ref{theorem2}. Note that EERM defines the entire graph sharing the same environment, while we refine the environment, considering the more common and challenging scenario where nodes come from multiple latent environments.

\begin{theorem}
\label{theorem2}
Under assumption \ref{assumption1} and \ref{assumption5}, if the invariant feature encoder \(\Phi\) and classifier \(c\) satisfy condition \ref{condition1} and \ref{condition3}, then the model \(f=c\circ \Phi\) is a solution to the OOD problem in Eqn. (\ref{equation2}), where the invariance condition is equivalent to \(I(Y;w(u(G))|\Phi(G))=0\), and the sufficiency condition is equivalent to maximizing \(I(Y;\Phi(G))\). The complete theoretical proof is provided in Appendix \ref{C}.
\end{theorem}

\subsubsection{Under what circumstances can IENE identify invariant features}
\label{Section3.3.2}
\begin{figure}[htbp] 
    \centering 
    \subfloat[ ]{%
        \includegraphics[width=0.1\textwidth]{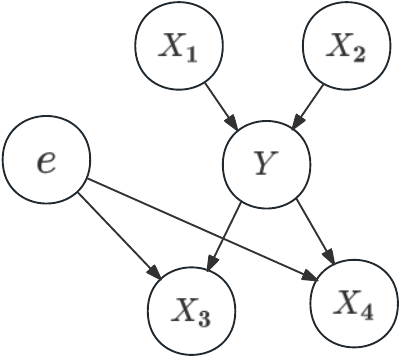}%
    }
    \quad 
    \subfloat[ ] {%
        \includegraphics[width=0.1\textwidth]{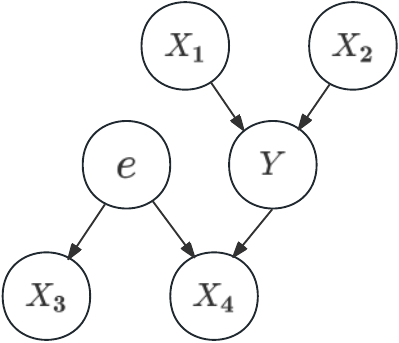}%
    }
    \quad 
    \subfloat[ ]{%
        \includegraphics[width=0.085\textwidth]{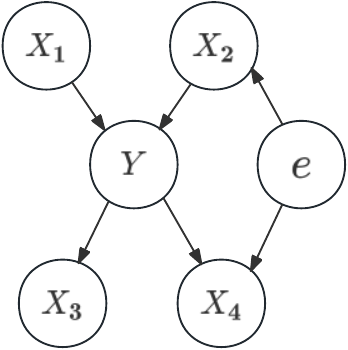}%
    }
    \caption{An example of causal structural model with certain general structural patterns, where \(X_1,X_2\) are invariant features, \(X_3,X_4\) are spurious features, \(Y\) is the target, and \(e\) is the environmental factor.}\label{figure1} 
\end{figure}

Figure \ref{figure1}. depicts an example of a causal structural model with certain general structural patterns, illustrating several scenarios where IENE can successfully identify invariant features without providing any prior knowledge of domain partitions. In Figure \ref{figure1}(a), we approximate the environmental factor \(e\) in the causal graph based on the inferred environmental partitions \(\rho\) by \(w(u(G))\). Then, IENE eliminates the influence of \(X_3,X_4\) through the invariance penalty (see Eqn. (\ref{equation4})), thereby identifying the invariant features \(X_1,X_2\). Figure \ref{figure1}(b) extends Figure \ref{figure1}(a), where we simultaneously identify the irrelevant feature \(X_3\) and the spurious feature \(X_4\) through environmental partitioning. Figures \ref{figure1}(a) and \ref{figure1}(b) can be extended to more complex scenarios, where, for simplicity, we omit representations of hidden confounding factors and other similar features. The case in Figure \ref{figure1}(c) is relatively complex because, in addition to spurious and irrelevant features, environmental partitioning may also penalize some invariant features (such as \(X_2\) in Figure \ref{figure1}(c)), as the assumption that \(P(Y|X_{i})\) remains invariant across different environments is not contradictory to \(X_2\gets  e \to X_4\). Therefore, IENE may unavoidably overlook a small portion of invariant features in such cases. However, this does not affect the ability of IENE to identify more influential and stable invariant features for predicting \(Y\). Further discussion is provided in Appendix \ref{B5}.

\subsection{Algorithm}
\label{section3.5}
In this section, we provide a concrete implementation of the concepts introduced in Section \ref{section3.2} and \ref{section3.3}. The complete algorithm is presented in Appendix \ref{A}.
\subsubsection{Identifying Invariant Features through Environment Partitioning}
\label{section3.5.1}
According to the analysis in Section \ref{section3.2}, to achieve environment partitioning, the first step is to extract environment representations \(h_{v_{e}}\) that satisfy \(h_{v_{e}}\perp h_{v_{i}}\) and \(X_v=d(h_{v_{e}}, h_{v_{i}})\) for each node \(v\). We developed a feature disentanglement framework to learn a feature extractor \(u\) and a feature reconstructor \(d\) that satisfy the conditions. Additionally, as \(h_{v_{i}}=\Phi(G_v)\) needs to be part of the input for \(d\), we should learn \(h_{v_{e}}\) and \(h_{v_{i}}\) simultaneously. Inspired by \textcite{Pan}, from the perspective of Disentangled Information Bottleneck, we can train multiple modules in an adversarial manner to obtain disentangled features: 
{
\abovedisplayskip=6pt 
\belowdisplayskip=6pt 
\begin{equation}
\begin{aligned}
\label{equation9}
&\min_{c,\Phi}\max_{w,c_1,...,c_K}\mathcal{R}(c,\Phi)
+\lambda \cdot 
{\textstyle \sum_{k=1}^{K}}\left [  
 \mathcal{R}^k_{\rho_{u,w}} (c,\Phi)
 -\mathcal{R}^k_{\rho_{u,w}} (c_k,\Phi)
 \right ]  \\&
 s.t. ~~u=\mathop{\text{argmin}}\limits_{u} D(d(u(G),\Phi(G)),X)
 +\eta \cdot 
\frac{\scalebox{.7}{HSIC}(u(G),\Phi(G))}{\sqrt{\scalebox{.7}{HSIC}(u(G),u(G)),\scalebox{.7}{HSIC}(\Phi(G),\Phi(G))} }
\end{aligned}
\end{equation}
}
where \(D\) represents the reconstruction loss, such as Mean Squared Error (MSE). \(u\) and \(w\) serve as graph neural networks since they need to learn representations from graph data \(G_v\), while \(d\), \(c\), and \(w\) can be constructed using typical multi-layer perceptrons. HSIC refers to the Hilbert-Schmidt Independence Criterion, and following \textcite{Kornblith,Chen04}, we use its empirical estimate and the normalized form. This learning process ensures that \(h_{v_{e}}\) and \(h_{v_{i}}\) are as mutually independent as possible, and \(h_{v_{e}}\) retains sufficient information from \(G_v\), extracting knowledge from irrelevant features and spurious features. Additionally, by imposing a penalty when \(\Phi\) extracts spurious features, as given by Eqn. (\ref{equation4}), \(\Phi\) gradually identifies more invariant features. This satisfies the conditions outlined in Section \ref{B6}.

\subsubsection{Identifying Invariant Features through Environment Extrapolation}
\label{section3.5.2}
According to the analysis in Section \ref{section3.3}, we leverage graph data augmentation to achieve environment extrapolation, identifying invariant features that are insensitive to structural changes across multiple augmented views. It's important to note that the environment recognizer is pre-trained using Eqn. (\ref{equation5}) at this point. Combining Eqn. (\ref{equation6}) and (\ref{equation8}), our objective is as follows:
{
\abovedisplayskip=6pt 
\belowdisplayskip=6pt 
\begin{equation}
\begin{aligned}
\label{equation10}
&\min_{c,\Phi}L_d=\mathcal{L}(\hat G^{(k)},Y)+\beta \cdot \mathbb{V}(\mathcal{L}(\hat G^{(k)},Y) )\\&
s.t.\ G_v^{(k)}= \mathop{\text{argmin}}\limits_{G_v^{(k)}}
\mathbb{E}_v  \left[   \mathbb{E}_k 
 \left[l(w(u({G}_v^{(k)} )),e^{(k)})  \right ] \right]
\end{aligned}
\end{equation}
}
where \(\hat G^{(k)}=(X,\hat{A}^{(k)})\), and \(\hat{A^{(k)}}=A+(E-I-2A)\circ S^{k}\), \(E\) is a matrix with all elements being 1, and \(I\) is an identity matrix. \( S^{k} \in \{0,1\}^{N\times N}\), where \(s^{k}_{ij} \in S^{k}\) represents whether the connection between node \(v_i\) and node \(v_j\) will be changed. As \(A\) is discrete and its gradient cannot be directly computed, following \textcite{Xu02}, we can sample \(S^k\) from the gradient of a surrogate loss using convex relaxation.

We avoid directly adjusting the features because, from causal perspective, changing invariant features without altering the target would modify the distribution \(P(Y|X_{i})\), contradicting the invariance assumption \ref{assumption1} and leading to the failure of identifying invariance. Due to the structural information of the graph, we can adjust the environment of the graph by modifying its structure without manipulating the original features. In essence, at this stage, we identify those invariant features that can still predict the target \(Y\) reliably under structural shifts. 

Ultimately, we integrate the environment partitioning with the environment extrapolation algorithm into a unified framework, as shown in Appendix \ref{A}. The advantages of our design are discussed in the analysis in Appendix \ref{D.5}. 

\section{Experiments}
\label{section4}
In this section, we evaluate the proposed methods using two synthetic and four real-world OOD datasets. ERM is chosen as the baseline method to demonstrate typical OOD performance under the IID assumption. We select EERM \cite{Wu02}, TIVA \cite{Tan}, CIE \cite{Chen04}, SRGNN \cite{Zhu}, LISA \cite{Lisa} and CaNet \cite{QTWu} for comparison with our method. Due to space limitations, we present the comparison with SRGNN and LISA in Appendix \ref{D}. Besides, Additional ablation experiments are presented in Appendix \ref{D}, and experimental details including  complete hyperparameter settings and the code of IENE are provided in Appendix \ref{E}.

\subsection{Out-of-Distribution Datasets}
Dataset Statistics. The statistical details of the datasets are presented in Table \ref{table1}, which includes three distinct types of distribution shifts: (1) Artificial Transformation, indicating that node features are replaced by synthetically generated spurious features; (2) Cross-domain Transfer, where the graphs in the dataset are from different domains; (3) Temporal Evolution, where the dataset consists of a dynamic graph with evolving properties. It is noteworthy that we utilize datasets provided by \textcite{Wu02} for OOD generalization assessment, involving distribution shifts crafted both from the aforementioned references and manually created. It is important to note that there can be multiple training/validation/testing graphs. Specifically, Cora \cite{Yang02} and Amazon-Photo \cite{Shchur} have 1/1/8 graphs for training/validation/testing sets. Similarly, Twitch-E \cite{Rozemberczki} has 11/1/5, FB-100 \cite{Traud} has 3/2/3, Elliptic \cite{Pareja} has 5/5/33, and OGB-Arxiv \cite{Hu} has 1/1/3.

\begin{table}[t]
\scriptsize
\centering
\caption{ Dataset Statistics. }
\label{table1}
\resizebox{1.0\columnwidth}{!}{
\begin{tabular}{cccccccc}
\hline
{Distribution Shift}&{Dataset}&{Nodes}&{Edges}&{Classes}&{Train/Val/Test Split}&{Metric}&{Adapted From}\\ 
\hline
\multirow{2}{*}{Artificial Transformation}{\centering}
& Cora& 2,703& 5,278& 10& Domain-Level& Accuracy& \textcite{Yang02}\\
& Amz-Photo& 7,650& 119,081& 10& Domain-Level& Accuracy& \textcite{Shchur}\\
\hline
\multirow{2}{*}{Cross-Domain Transfers}
& Twitch-E& 1,9129,498& 31,299-153,138& 2& Domain-Level& ROC-AUC& \textcite{Rozemberczki}\\
& FB100& 76941,536& 16,656-1,590,655& 2& Domain-Level& Accuracy& \textcite{Traud}\\
\hline
\multirow{2}{*}{Temporal Evolution}
& Elliptic& 203,769& 234,355& 2& Time-Aware& F1 Score& \textcite{Pareja}\\
& OGB-Arxiv& 169,343& 1,166,24& 40& Time-Aware& Accuracy& \textcite{Hu}\\
\hline
\end{tabular}
}
\end{table}

\subsection{Generalization to Out-Of-Distribution Data}
\label{section4.2}
IENE-r represents the integration of environment partitioning and invariant learning methods (Section \ref{section3.2}), while IENE-e denotes the combination of environment extrapolation and invariant learning methods (Section \ref{section3.3}). IENE-re signifies the integration of both approaches. As these algorithms can be executed independently, we evaluate their individual performances, as detailed in Appendix \ref{D.5}.

Here, we compare IENE-re with five methods: Empirical Risk Minimization (ERM, i.e., standard training), EERM \cite{Wu02}, CIE \cite{Chen04}, TIVA \cite{Tan}, and CaNet \cite{QTWu}. Additionally, we evaluate all methods using four popular GNN backbones, including GCN \cite{Kipf}, GraphSAGE \cite{Hamilton}, GAT \cite{Velickovic}, and GPR \cite{Chien}. The parameter settings for the compared methods follow the original papers, and we performed parameter tuning on different datasets. For more implementation details of the baselines and IENE, please refer to Appendix \ref{E}. It is worth noting that all experiments in this paper were repeated 10 times with different random seeds.

 
\begin{table}[t]
\scriptsize
   \centering %
   \caption{ Average classification performance (\%) on test graphs. "Rank" indicates the average rank for each method across different backbone networks. "OOM" denotes an out-of-memory error on a GPU(A40) with 48 GB memory. Compared to the baselines, the proposed IENE-re consistently achieves the highest ranking. }
   \label{table2}
   \vskip 0.0in
  	\resizebox{1.0\columnwidth}{!}
{      
\begin{tabular}{ccccccccc}
\hline

Backbone&  Method& Cora& Amz-Photo& Twitch-E& FB-100& Elliptic& OGB-Arxiv& Rank\\ 
\hline
\multirow{5}{*}{GCN}
& ERM& 92.29$\pm$ 3.87& 92.34$\pm$0.55& 54.04$\pm$ 6.46& 52.52$\pm$0.61& 61.86$\pm$1.29& 41.59$\pm$1.20& 5.5\\
& EERM& 92.49$\pm$ 1.55& 91.54$\pm$0.45& 56.28$\pm$ 5.93& 53.59$\pm$0.12& 62.35$\pm$3.57& 41.76$\pm$1.07& 4.3\\
& CIE& 93.17$\pm$ 0.92& 93.59$\pm$1.13& 59.15$\pm$ 3.61& 52.92$\pm$1.21& 65.13$\pm$2.83& 40.88$\pm$1.30& 4.0\\
& TIVA& 94.49$\pm$ 1.35& 95.61$\pm$3.18& 55.15$\pm$ 1.58& 52.33$\pm$0.87& 64.29$\pm$2.14& 43.74$\pm$0.61& 4.0\\
& CaNet& \textbf{95.80$\pm$ 1.04}& 95.93$\pm$0.55& 61.47$\pm$ 0.32& 53.30$\pm$1.83& 66.45$\pm$2.01& \textbf{45.19$\pm$0.74}& 1.8\\
& IENE-re& 95.58$\pm$ 2.07& \textbf{96.40$\pm$1.80}& \textbf{61.87$\pm$ 1.43}& \textbf{53.73$\pm$2.65}& \textbf{67.90$\pm$3.31}& 44.85$\pm$0.31& \textbf{1.3}\\
\hline
\multirow{5}{*}{SAGE}
& ERM& 98.02$\pm$ 0.63& 95.03$\pm$1.02& 65.14$\pm$ 0.52& OOM& 58.08$\pm$3.80& 40.37$\pm$0.83& 5.4\\
& EERM& \textbf{99.58$\pm$ 0.03}& 96.55$\pm$0.94& 66.69$\pm$ 0.26& OOM& 61.18$\pm$4.77& 41.00$\pm$1.16& 2.6\\
& CIE& 98.20$\pm$ 0.38& 94.59$\pm$1.27& 65.07$\pm$ 0.44& OOM& 59.91$\pm$2.59& 40.45$\pm$0.99& 5.2\\
& TIVA& 99.13$\pm$ 0.89& 94.30$\pm$1.93& 65.76$\pm$ 0.16& OOM& 63.74$\pm$4.10& 41.59$\pm$1.71& 3.8\\
& CaNet& 99.41$\pm$ 0.27& 96.13$\pm$1.14& \textbf{66.84$\pm$ 0.61}& OOM& 67.95$\pm$2.93& 41.53$\pm$1.08& 2.2\\
& IENE-re& 99.27$\pm$ 0.55& \textbf{96.87$\pm$1.65}& 66.18$\pm$ 0.78& OOM& \textbf{68.38$\pm$3.78}& \textbf{41.82$\pm$1.84}& \textbf{1.8}\\
\hline
\multirow{5}{*}{GAT}
& ERM& 91.26$\pm$ 3.26& 94.81$\pm$1.06& 58.86$\pm$ 2.69& 53.22$\pm$0.34& 64.65$\pm$3.68& 43.37$\pm$1.10& 5.8\\
& EERM& 96.19$\pm$ 2.48& 95.32$\pm$0.73& 62.43$\pm$ 1.30& 53.31$\pm$0.29& 62.04$\pm$2.19& 44.05$\pm$4.04& 4.2\\
& CIE& 94.31$\pm$ 1.52& 95.07$\pm$0.35& 61.02$\pm$ 1.81& 53.68$\pm$0.82& 64.95$\pm$3.21& 44.54$\pm$1.07& 4.0\\
& TIVA& 97.50$\pm$ 0.98& 96.51$\pm$1.40& 60.74$\pm$ 1.85& 53.49$\pm$1.92& 65.38$\pm$2.17& 43.80$\pm$1.21& 3.8\\
& CaNet& 98.12$\pm$ 0.74& 96.60$\pm$0.95& 61.87$\pm$ 1.69& 54.38$\pm$1.77& 66.12$\pm$2.31& 45.49$\pm$1.40& 2.2\\
& IENE-re& \textbf{99.00$\pm$ 0.60}& \textbf{96.85$\pm$1.76}& \textbf{62.55$\pm$ 1.42}& \textbf{54.72$\pm$ 2.05}& \textbf{66.56$\pm$3.95}& \textbf{45.70$\pm$0.81}& \textbf{1.0}\\
\hline
\multirow{5}{*}{GPR}
& ERM& 84.94$\pm$ 2.87& 91.35$\pm$0.81& 63.27$\pm$ 0.34& 54.10$\pm$0.72& 62.82$\pm$2.23& 47.16$\pm$0.52& 4.8\\
& EERM& 88.85$\pm$1.59& 89.27$\pm$3.61& \textbf{65.72$\pm$ 0.71}& 54.86$\pm$0.58& 61.00$\pm$0.39& 45.93$\pm$1.12& 4.5\\
& CIE& 90.11$\pm$ 1.70& 91.86$\pm$1.59& 63.53$\pm$ 0.18& 54.45$\pm$1.32& 62.63$\pm$2.91& 46.21$\pm$0.95& 4.3\\
& TIVA& 92.12$\pm$ 2.19& 92.86$\pm$1.46& 64.46$\pm$ 0.63& 54.21$\pm$0.87&62.11$\pm$0.48& 47.09$\pm$1.53& 3.8\\
& CaNet& 93.64$\pm$ 2.51& 92.58$\pm$2.39& 65.33$\pm$ 0.45& 55.61$\pm$0.94&63.18$\pm$1.08& 47.35$\pm$1.37& 2.2\\
& IENE-re& \textbf{94.36$\pm$ 3.34}& \textbf{92.93$\pm$2.76}& 64.68$\pm$ 0.72& \textbf{55.75$\pm$ 1.28}& \textbf{63.57$\pm$1.65}& \textbf{47.53$\pm$1.20}& \textbf{1.3}\\
\hline
\end{tabular} 
}                 
\end{table}

The results in Table \ref{table2} report the average performance on test graphs for each dataset and the average ranking for each algorithm. From the table, we make the following observations:

(a) Overall Performance: The framework consistently demonstrates optimal performance on all OOD datasets, with IENE achieving average rankings of 1.3, 1.8, 1.0, and 1.3 when using GCN, SAGE, GAT, and GPR as backbone methods, respectively. Furthermore, in most cases, IENE significantly outperforms the vanilla baseline (ERM). Notably, when using GCN as the backbone, IENE outperforms ERM by 3.29\%, 7.83\%, and 6.04\% on Cora, Twitch-E, and Elliptic, respectively. These results highlight the superiority of IENE in addressing different types of distribution shifts.

\begin{wrapfigure}{r}[0cm]{0pt}        
    \includegraphics[width=5cm]{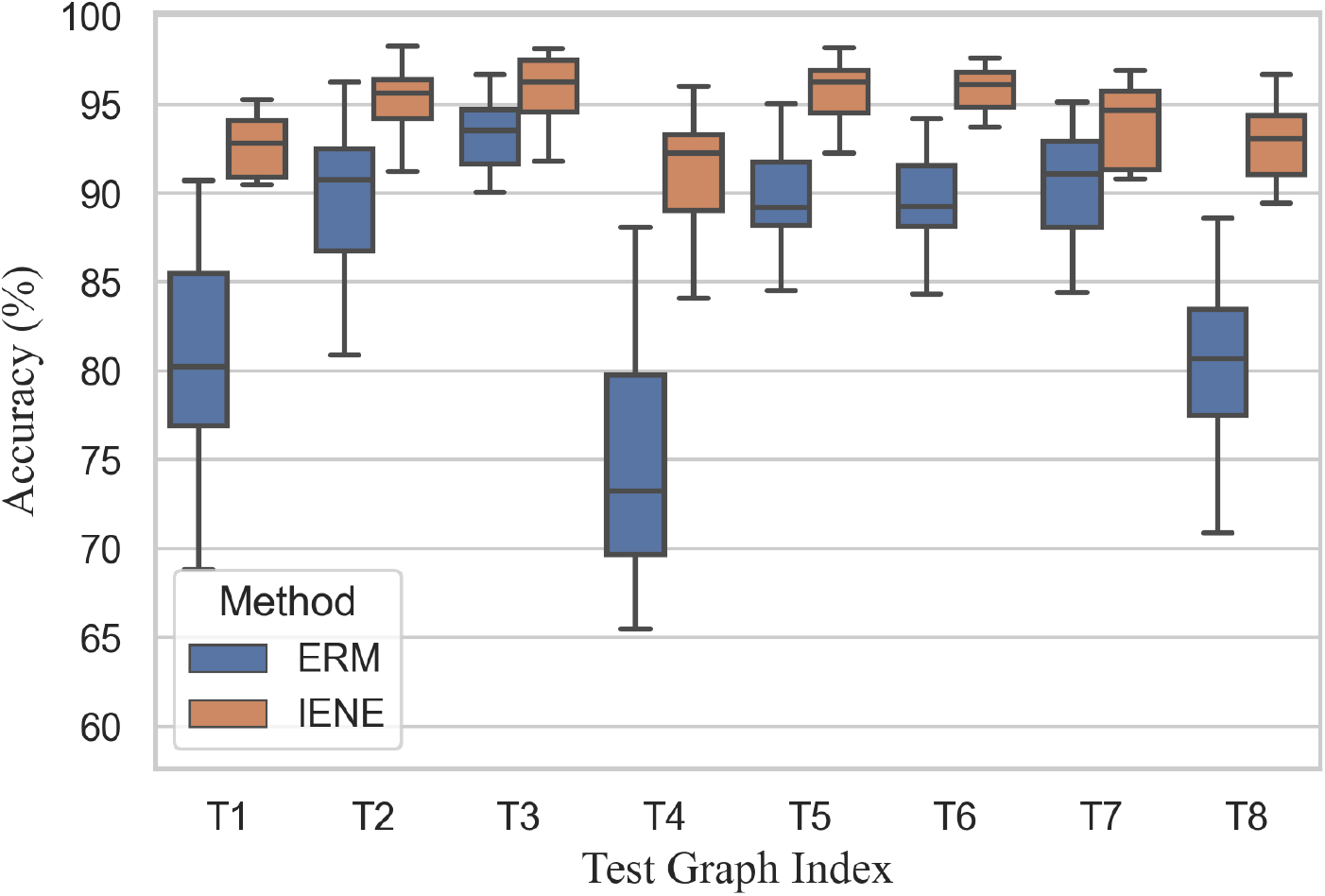} 
    \caption{Results on Cora under OOD. Compared to ERM, IENE significantly improves the accuracy and stability of GNN.} 
    \label{figure2}
\end{wrapfigure}

(b) Comparison with Other Baselines. CIE, TIVA, CaNet and EERM all modify the training process to enhance the generalization ability of the model. However, in most cases, these methods demonstrated suboptimal performance. As a training approach for automatically identifying environments, TIVA demonstrated satisfactory performance in certain scenarios, but it is limited to addressing OOD problems on datasets that conform to its assumptions ($X_z\perp Y$ and $X_z$ has clear environmental information). We observe that IENE generally outperforms other methods in tests on each dataset. Furthermore, we present an additional analysis of the performance on each test graph using a model trained on Cora in Figure \ref{figure2}.

(c) Efficiency Comparison. Table \ref{table3} illustrates the efficiency comparison of the three variants of IENE, EERM, and TIVA on the Cora dataset. The backbone network for experiments is GCN, and the hyperparameters for these frameworks are unified. Note that, due to the two-stage training strategy employed in both IENE and TIVA, the runtime for every 100 epochs in the table represents the average time for both stages. The time cost of IENE-r is mainly determined by the bi-level optimization between the classifiers in discrete environments and the shared classifier, along with the alternating learning of the two representation extractors. The time cost of IENE-e primarily depends on the bi-level optimization between the view augmentation component and the invariant learner. Compared to the retraining method EERM, IENE-e adopts a similar training pattern but exhibits higher efficiency. IENE-r and TIVA share a similar training pattern, resulting in a small efficiency gap between them. Furthermore, IENE-re combines the two proposed methods, achieving superior OOD generalization performance without incurring significant additional memory overhead. Note that both IENE-e and IENE-re include the time and memory occupied by the pre-training phase of IENE-r.

\section{Conclusion}
\label{section5}
In this paper, we propose a novel method, IENE, to achieve OOD generalization of GNNs from the perspective of invariant learning. First, we present a fully data-driven environmental inference method that extracts environmental representation and partitions environments with a disentangled information bottleneck framework, simultaneously discovers invariance in features. Besides, we analyze the characteristics of invariance on graph-structured data and design a augmentation-based method for environmental extrapolation, generating a series of environmental views to learn invariance in the structure of graphs. Additionally, we integrate environmental identification, environmental extrapolation techniques, and an invariance learning framework to discover invariant patterns on graph. In practice, we have implemented conceptual methods specifically, and for stability, we conduct invariant learning in a bilevel optimization manner. We theoretically and experimentally demonstrate that IENE can automatically learn invariance from input graphs, achieving competitive generalization performance on OOD datasets. The evaluation also validated our theoretical analysis. In future work, We will enhance the scalability of IENE to adapt to large-scale graphs and apply the method to specific GNN-based real-world scenarios to address the OOD problem.

\begin{ack}
Use unnumbered first level headings for the acknowledgments. All acknowledgments
go at the end of the paper before the list of references. Moreover, you are required to declare
funding (financial activities supporting the submitted work) and competing interests (related financial activities outside the submitted work).
More information about this disclosure can be found at: \url{https://neurips.cc/Conferences/2024/PaperInformation/FundingDisclosure}.

Do {\bf not} include this section in the anonymized submission, only in the final paper. You can use the \texttt{ack} environment provided in the style file to automatically hide this section in the anonymized submission.
\end{ack}

\small 
\printbibliography
\normalsize 

\newpage
\appendix

\onecolumn
\section*{Appendix}
\label{Appendix}
\section{Learning Algorithm}
\label{A}
IENE-re combines IENE-r and IENE-e, and its objective function is given by Eqn. (\ref{equation11}).
{
\abovedisplayskip=6pt 
\belowdisplayskip=6pt 
\begin{equation}
\begin{aligned}
\label{equation11}
&\min_{c,\Phi}\max_{w,c_1,...,c_K}\mathcal{R}(c,\Phi)
+\lambda \cdot 
{\textstyle \sum_{k=1}^{K}}\left [  
 \mathcal{R}^k_{\rho_{u,w}} (c,\Phi)
 -\mathcal{R}^k_{\rho_{u,w}} (c_k,\Phi)
 \right ]  
 +\mathcal{L}(\hat G^{(k)},Y)+\beta \cdot \mathbb{V}(\mathcal{L}(\hat G^{(k)},Y) )
\end{aligned}
\end{equation}
}
Detailed variable explanations can be found in Eqn. (\ref{equation9}) and (\ref{equation10}). Since the Eqn. (\ref{equation11}) involve a max-min objective, it is complex and challenging to optimize directly. Therefore, in practice, we can complete the learning of different components in stages. In general, during the implementation process, we can follow a two-stage training approach to optimize the model. The complete algorithm is provided in \cref{alg:example}.

\begin{algorithm}[tb]
   \caption{IENE-re}
   \label{alg:example}
\begin{algorithmic}
   \STATE {\bfseries Input:} invariant feature extractor $\Phi$, invariant label classifier $c$, 
   environmental feature extractor $u$, environmental index classifier $w$, environmental label classifier $c_1,...,c_K$, feature reconstructor $d$, Input graph $G$ and target $Y$.
   \STATE \textbf{1.Stage-one: Annealing Iteration}
   \FOR{each annealing iteration}
   \FOR{each disentanglement iteration}
   \STATE Train $u, d$ by minimize the constraint part of Eqn. (\ref{equation9}) with fixed $\Phi$.
   \ENDFOR
   \STATE Train \(c_1,...,c_K\) by minimizing the ERM loss \( {\textstyle \sum_{k=1}^{K}}\mathcal{R}(c_k,\Phi)\) with fixed $\Phi$.
    \STATE Train $\Phi, c$,  by minimize the Eqn. (\ref{equation5}) with fixed $u, w$ and \(c_1,...,c_K\).
    \STATE Train $w$ by maximize the Eqn. (\ref{equation5}) with fixed $u, \Phi, c$ and \(c_1,...,c_K\).
   \ENDFOR
    \STATE \textbf{2.Stage-two: Invariant Learning Iteration}
    \FOR{each training iteration}
   \FOR{each structural intervention iteration}
   \STATE Train $\hat{G}_v^{(1)},...,\hat{G}_v^{(K)}$ by minimize the Eqn. (\ref{equation6}) with fixed $u, w$.
   \ENDFOR
    \STATE Train $\Phi, c$ by minimize the Eqn. (\ref{equation11}) with fixed $u, w$ and \(c_1,...,c_K\).
   \ENDFOR
\end{algorithmic}
\end{algorithm}

\section{Conceptual Analysis and Discussion}
\subsection{Auxiliary Information for Identifying the Environment}
\label{B1}
$X_{v_{i}}$ provides no benefits for environment identification and may even introduce additional noise, while $X_{v_{s}}$ and $X_{v_{ir}}$ play an important role in inferring the environment. According to the assumption, the probability distribution $P(Y|X_{v_{s}})$ varies across different environments $e\in \varepsilon$, while $P(Y|X_{v_{i}})$ remains invariant, i.e., $P(Y|X_{v_{i}},e)=P(Y|X_{v_{i}},e')$.

Here, we consider two cases of $X_{v_{i}}$.

The first case corresponds to Figure \ref{figure1}(a)(b), where the nodes in the figure are generated based on a Structural Causal Model (SCM) with a directed acyclic graph. Assuming the Markov property and other underlying assumptions hold, in both causal graphs, we have Condition $X_{i}\bot e$, indicating that $X_{i}$ manifests as noise relative to the environment $e$. 

The second case corresponds to Figure \ref{figure1}(c). Clearly, in this causal graph, $X_{i}\not \!\bot\ e$, due to $e\to X_2$. While $X_2$ may contribute to environmental inference, it is essential to emphasize that within the framework of invariant learning, the purpose of environment partitioning is to maximize the differences in $P(Y|X_{v_{s}})$ across different partitions, providing an invariant penalty. This aims to prevent the invariant representation learner $\Phi$ from extracting spurious features $X_{v_{s}}$ from the input graph. If we use $X_2$ as the input for environmental prediction, where different values $X_2^{(1)}\neq X_2^{(2)}$ of $X_2$ correspond to inferred environments $e^{(1)}$ and $e^{(2)}$, we have $P(Y|X_2,e^{(1)})\neq P(Y|X_2,e^{(2)})$. Consequently, the subset of invariant features $X_2\in X_{i}$ is considered by the invariant learner as spurious, hindering the learning of invariant features with similar properties. 

Therefore, $X_{i}$ should not be used for predicting the environment in any case.

According to the previous assumptions, $P(Y|X_{v_{s}},e)\neq P(Y|Y|X_{v_{s}},e')$. In the causal graph, the relationship between $X_{v_{s}}$ and the environment $e$ can be considered as a direct or latent causal relationship. Therefore, it is reasonable to use $X_{v_{s}}$ as a basis for predicting the environment. Literature \cite{Tan} assumes that features $X_{v_{ir}}$ unrelated to the target $Y$ can be used to predict the environment, such as background factors like time or location. While this assumption is possible, it may not always hold. Thus, we believe that both $X_{v_{s}}$ and $X_{v_{ir}}$ play important roles in environmental reasoning.

\subsection{Adaptation to Structural Environment Changes}
\label{B2}
Training an invariant model on static graph data may face challenges in adapting to changes in the structural environment. In data with independent samples, Method in \ref{section3.2} theoretically excels at identifying invariant features. However, in graph-structured data, nodes’ representations are not independent of those of their neighbors. This characteristic leads the learner $\Phi$ (composed of some GNN layers, to learn high-quality information from graph-organized data), to inevitably aggregate the invariant features of the neighbors when learning the invariant representation $h_{v_{i}}$ of node $v$. When structural changes occur, this aggregation may significantly impact the learned invariant representation, resulting in failures in downstream tasks. In other words, training $\Phi$ using Eqn. (\ref{equation5}) can only guarantee the identification of invariant features on graphs with similar structural properties. For a single graph, Method in \ref{section3.2} can only learn to identify invariant features under a static context, and it is not invariant to changes in structural properties. Therefore, we extend the method to environment extrapolation, conducting dynamic invariant learning through environment augmentation. This is verified in the ablation experiments in Appendix \ref{D.5}.

\subsection{The Limitations of V-REx}
\label{B3}
In certain cases of node classification, V-REx \cite{Krueger} may struggle to impose a significant penalty for dissimilarity. Here, we illustrate this issue with a simple example. We consider two views, $G^{(1)}$ and $G^{(2)}$, from different environments. When $\mathbb{E}_vl(f(G_v^{(1)}),y_v)\approx \mathbb{E}_vl(f(G_v^{(2)}),y_v)$, V-REx $\mathbb{V}_k(\mathcal{L}((G^{(1)}),Y))\approx 0$, but for a node $v\in V$, there may be a significant difference between $l(f(G_v^{(1)}),y_v)$ and $l(f(G_v^{(2)}),y_v)$, indicating that graph-level V-REx fails to capture node-level differences. In more extreme cases, where nodes correctly classified in one view are complementary to those in the other view, they may still exhibit similar loss values at the graph level (However, from the perspective of node-wise environment, these two views are vastly different). Therefore, V-REx perceives their performance as similar and may not impose a significant penalty. NV-REx, on the other hand, is capable of providing a positive penalty based on the differences of each node in different environments, accumulating them to better penalize the learned features for their dissimilarity across environments, compelling the learner to acquire invariant features.

\subsection{Feature Learning Preserves Invariance}
\label{B4}
In IENE, the invariant representation $h_{v_{i}}=\Phi (G_v)$ is capable of preserving observations of $X_{v_{i}}$. In this section, following \textcite{Arjovsky,Rosenfeld,Lin01}, we will demonstrate that IENE can learn invariant features given a scrambled observation in a linear form. Specifically, we consider the same data generation process as \textcite{Arjovsky}: 
{
\abovedisplayskip=6pt 
\belowdisplayskip=6pt 
\begin{equation}
\begin{aligned}
\label{equation12}
&Y^e=X_{i}^e\cdot \beta +\epsilon _{i}, X_{i}^e\perp \epsilon _{i}, 
\mathbb{E}(\epsilon _{i})=0, X^e=W\cdot [X_{i}^e,X_{s}^e]
\end{aligned}
\end{equation}
}
where $\beta\in \mathbb{R}^{d_{i}}$ and $W\in \mathbb{R}^{d\times (d_{i}+d_{s})}$. We assume the existence of $\widetilde{W}\in \mathbb{R}^{d_{i}\times d}$ such that $\widetilde{W}(W[x_{i};x_{s}])=x_{i}$ for all $x_{i}$ and $x_{s}$. Both the feature extractor and predictor adopt a linear form, that is, $\Phi$ takes values in $\mathbb{R}^{d\times d}$ and $x_{i}$ takes values in $\mathbb{R}^d$. The prediction for $X$ is $\omega \circ  \Phi(X)=(\Phi X)^T\omega$.

In this case, a major challenge lies in describing the impact of an invariant or spurious feature in a quantitative manner: the feature extractor $\Phi$ can capture arbitrary small portions of spurious information. Following \textcite{Arjovsky,Rosenfeld,Lin01}, we consider a constrained form of problem \ref{equation5} for theoretical analysis:
{
\abovedisplayskip=6pt 
\belowdisplayskip=6pt 
\begin{equation}
\begin{aligned}
\label{equation13}
&\min_{c,\Phi}\mathcal{R}(c,\Phi), s.t. \max_{u,w,c_k} {\textstyle \sum_{k=1}^{K}}[
\mathcal{R}^k_{\rho _{u,w}}(c,\Phi)- \mathcal{R}^k_{\rho _{u,w}}(c_k,\Phi)
] 
\end{aligned}
\end{equation}
}
Similar to condition \ref{condition1} and condition \ref{condition2}, auxiliary information should also provide sufficient information so that the inferred environment can be diverse enough while maintaining latent invariance. This aligns with existing conditions for identifiability in linear cases \textcite{Rosenfeld,Lin01}. In this paper, we leverage such a condition, $linear general position condition$, from \textcite{Arjovsky}. For our analysis, we use squared error as the loss function and consider that $\rho(\cdot)$ partitions the  environments in a challenging way, where each data sample is precisely assigned to one environment. We use $L(\cdot|\cdot)$ to denote the optimal expected risk. We also assume that environments are non-degenerate, meaning each inferred environment contains some data samples; otherwise, we can simply remove such environments. The identifiability results for the linear case are as follows.
\begin{proposition}
\label{proposition1}
Suppose $H(\cdot|\cdot)$ in condition \ref{condition1} is replaced by $L(\cdot|\cdot)$. Assume the existence of $\rho(\cdot)$ such that the generated environments, denoted as $\{X^k\}_{k=1}^K$, are in linear generic position of degree $r$. For example, for some $r\in \mathbb{N}$, there exists $K>d-r+d/r$, and for any non-zero $x\in \mathbb{R}^d:dim(span(\{\mathbb{E}_{X^k}[X^kX^{k^T}]x-\mathbb{E}_{X^k,\epsilon _{i}}[X^k]\epsilon _{i}\}_k))>d-r$. If the rank of $\Phi \in \mathbb{R}^{d\times d}$, $r>0$, then Eqn. (\ref{equation13}) will yield the expected invariant predictor.
\end{proposition}

\textit{Proof.} \textbf{Step 1}: No spurious feature will be learned. Given a partition $\{X^k\}_{k=1}^K$ in linear generic position of degree $r$, theorem 9 in \textcite{Arjovsky} demonstrates that for any $k$, $\Phi$, and $\omega$ satisfying the normal equations $\Phi \mathbb{E}_{X^k}[X^kX^{k^T}]\Phi^T\omega =\Phi\mathbb{E}_{X^k,Y^k}[X^kY^k]$, it holds if and only if $\Phi$ leads to the desired invariant predictor $\Phi^T\omega=\widetilde{W} ^T\beta$. Therefore, we only need to prove that our solution satisfies the same normal equations. Let $\Phi'$ and $\omega'$ denote the solution to Eqn. (\ref{equation13}). According to the constraints and the general linear position condition, we know that there exists a partition $\{X^k\}_{k=1}^K$ in general linear position of degree $r$, where $\mathcal{R}^k(\omega',\Phi')=\mathcal{R}^k(\omega_k,\Phi')$. Note that $\omega_k$ only minimizes the mean squared error in the $k$-th environment, so it must satisfy the normal equation $\Phi' \mathbb{E}_{X^k}[X^kX^{k^T}]\Phi'^T\omega =\Phi'\mathbb{E}_{X^k,Y^k}[X^kY^k]$. As $\omega'$ achieves the same minimal mean squared error in the $k$-th environment, $\omega'$ must also satisfy the normal equations. Therefore, the invariant representation contains no spurious information, $X\Phi=[SX_{i};0_{d_{s}}]$, where $S\in \mathbb{R}^{d_{i}\times d_{i}}$ is a reversible matrix, and $0_{d_{s}}$ represents $d_{s}$- dimensional vectors with all elements being zero.

\textbf{Step 2}: The representation does not discard any invariance information. With $H(\cdot|\cdot)$ replaced by $L(\cdot|\cdot)$ in condition \ref{condition1}, we have $L(Y|X_{i})-L(Y|X_{i},w(u(G)))=0$. Then $X\Phi$ will satisfy the constraints of Eqn. (\ref{equation13}). Note that the obtained loss for $X\Phi$ is minimized when only using invariant feature information.

\subsection{More Discussions on Identifiable Invariance}
\label{B5}
IENE may inevitably overlook a small portion of invariant features, but this does not affect our ability to identify more influential and stable invariant features for predicting $Y$. Considering the case in Figure \ref{figure1}(c), $X_2$, a subset of invariant features may be mistaken as spurious features due to changes in the environment, leading to $P(Y|X_2,e^{(1)})\neq P(Y|X_2,e^{(2)})$. However, we can adjust the learning extent of the invariant learner for features like $X_2$ by controlling the trade-off between ERM loss and invariant penalty. When the invariant penalty dominates, the invariant learner tends to reject learning $X_2$ (while rejecting all spurious features, but sacrificing information in $X_2$ that is helpful for predicting $Y$); when the ERM loss dominates, it tends to learn from $X_2$ because it contains causal features for $Y$ (retaining most of the information helpful for predicting $Y$ in $X_2$, but possibly learning some spurious features $x_s\in X_4$). Invariant features $X_2^s \subseteq X_2$, which play a more critical role in predicting $Y$, are not overlooked, as this would significantly increase the ERM loss and dominate the training, prompting the relearning of invariant features from $X_2^s$. 

\subsection{Prerequisites and Assumptions}
\label{B6}
In IENE, the invariant representation \(h_{v_{i}}=\Phi(G_v)\) not only preserves observations of the original invariant features \(X_{v_{i}}\) in the raw data but also captures new invariances, as demonstrated in Appendix \ref{B4}. Based on this, we can relax the assumptions about the data.

\begin{assumption}
\label{assumption1}
There exist invariant features \(X_{v_{i}}\) in the original features \(X_{v}\) of \(G_{v}\), or \(h_{v_{i}}\) obtained by some linear transformation \(\Phi\) on the original features \(X_{v}\) satisfies the invariance property, i.e., Property \ref{Property1}. 
\end{assumption}

\begin{property}
\label{Property1}
(Invariance Property): In causal graph, under any intervention at any node (except \(Y\) itself), \(P(Y|X_{i})\) or \(P(Y|h_{i})\) remains invariant. 
\end{property}

For simplicity, in the subsequent descriptions, we use \(X_{i}\) to refer to \(X_{i}\) or \(h_{i}\), and \(X_{s}\) to refer to \(X_{s}\) or \(h_{s}\), as they share similar properties. We then introduce the following assumptions, which are similar to assumptions 1-3 in the literature \cite{Lin01,Tan}.

\begin{assumption}
\label{assumption2}
For any given learning function \(\Phi\) and any constant \(\epsilon>0\), there exists \(c\in F\), such that \(\mathbb{E}[\mathcal{L}(c(\Phi(G)),Y)]\le H(Y|\Phi(G))+\epsilon\). 
\end{assumption}

\begin{assumption}
\label{assumption3}
If a feature violates the invariance constraint, adding another feature will not eliminate the penalty, i.e., there exists a constant \(\delta>0\) such that for spurious feature \(X_1\subset X_{s}\) and any feature \(X_2\subset X\), we have \(H(Y|X_1,X_2)-H(Y|w(u(G)),X_1,X_2)\ge \delta(H(Y|X_1)-H(Y|w(u(G)),X_1))\).
\end{assumption}

\begin{assumption}
\label{assumption4}
Let \(X_{i}^s\) represent any appropriate subset of invariant features, i.e., \(X_{i}^s \subsetneq X_{i}\), we have \(H(Y|X_{i})\le H(Y|X_{i}^s)-\gamma \), where \(\gamma\) is a constant and \(\gamma>0\).
\end{assumption}

Assumption \ref{assumption2} is a common assumption, requiring that the function space is rich enough. So, given \(\Phi\), there exists \(c\in F\) that can fit \(P(Y|\Phi(G))\) well. The purpose of Assumption \ref{assumption3} is to ensure sufficient positive penalty when spurious features are included. Assumption \ref{assumption4} indicates that any set of features contains some information useful for predicting \(Y\). Otherwise, we could simply remove such a class of irrelevant features as they do not impact predictions.

As discussed in Section \ref{section3.4.2}, if \(u(\cdot)\), \(w(\cdot)\), \(\Phi(\cdot)\), and \(c(\cdot)\) satisfy condition \ref{condition1}, \ref{condition2}, and \ref{condition3}, Eqn. (\ref{equation5}) and (\ref{equation8}) can be proven to learn invariant features, and the downstream classifier exhibits the same and optimal performance across different environments.

\begin{condition}
\label{condition1}
(Invariance Condition): Given the invariant feature \(X_{i}\) and any environment index classifier \(w(\cdot)\), it holds that \(H(Y|X_{i},w(u(G)))=H(Y|X_{i})\).
\end{condition}
\begin{condition}
\label{condition2}
(Non-Invariance Discrimination Condition): For any feature \(X_{s}^s \in X_{s}\), there exists an environment index classifier \(w(\cdot)\) and a constant \(C>0\), such that \(H(Y|X_{s}^s)-H(Y|X_{s}^s,w(u(G)))\ge C\). 
\end{condition}
\begin{condition}
\label{condition3}
(Sufficiency Condition): For the invariant feature encoder \(\Phi(\cdot)\) and classifier \(c(\cdot)\), \(Y=c(\Phi(G))+n\), where $n$ is independent noise.
\end{condition}

Condition \ref{condition1} requires that the invariant feature \(X_{i}\) caused by \(w(u(G))\) should remain invariant for any environment partition. Otherwise, if there is a partition where an invariant feature changes, that feature will result in a positive penalty. Condition \ref{condition2} implies that, for each spurious feature, there is at least one partition where this feature is non-invariant in a split environment. If a spurious feature does not receive any invariance penalty in all possible environment partitions, we can never distinguish it from a truly invariant feature. 

Then, the next assumption will serve as the foundation for our environmental extrapolation approach.

\begin{assumption}
\label{assumption5}
(Environmental Heterogeneity): For $\Phi$, $u$, and $w$ satisfying conditions \ref{condition1} and \ref{condition2}, there exists a function mapping $m$ such that $G^{(e)}=m(G,n^{(e)})$. And $P(Y|X^k_{s},w(u(G)))$ can arbitrarily vary across different environments.
\end{assumption}

\textbf{Are these assumptions difficult to hold?} These assumptions are explicitly or implicitly required in a series of works such as IRM, ZIN, TIVA, and VREx. Therefore, we believe that assumptions 1-4 used in this paper are nearly unavoidable in the current framework of invariant learning based on environmental partitioning. And assumption 5 is also a common assumption in graph invariant learning works such as EERM and DIR. We believe that the success of these works indicates that these assumptions are generally easy to hold in reality. 

\section{Proofs}
\label{C}
\subsection{Proof of Theorem 1}
To prove Theorem \ref{theorem1}, we can replace $X_v$ and $X_s$ with $h_{v_{i}}$ and $h_{v_{s}}$ in the proof of Theorem 2 as presented by \textcite{Lin01}.

\subsection{Proof of Theorem 2}
\begin{lemma}
\label{lemma1}
Condition \ref{condition1} (Invariance) can be equivalently expressed as: $I(Y;w(u(G))|h_{i})=0$, and Condition \ref{condition3} (Sufficiency) can be equivalently represented as: $I(Y;h_{i})$ is maximized. 
\end{lemma}
\textit{Proof.} For invariance, we can easily derive equivalent expressions for the given facts.
{
\abovedisplayskip=6pt 
\belowdisplayskip=6pt 
\begin{equation}
\begin{aligned}
\label{equation14}
&I(Y;w(u(G))|h_{i})=D_{KL}(P(Y,w(u(G))|h_{i})||P(Y|h_{i})P(w(u(G))|h_{i}))\\
&=D_{KL}(P(Y|w(u(G)),h_{i})||P(Y|h_{i}))
] 
\end{aligned}
\end{equation}
}
For sufficiency, we first prove that if $Y=c^*(h_{i})+n$ is satisfied, then $h_{i}=argmax_hI(Y;h)$ will also be satisfied simultaneously. Let $h'=argmax_hI(Y;h), h_{i}\in supp(h_{i})$, and we prove this by showing that $I(Y;h')\le I(Y;h_{i})$. There exists a random variable $\overline{h}_{i}$ that is independent of $h_{i}$ and $h'$ can be expressed as $h'=m(h_{i},\overline{h}_{i})$, where $m$ is a mapping function. Then, we can derive the following:
{
\abovedisplayskip=6pt 
\belowdisplayskip=6pt 
\[
I(Y;h')=I(Y;m(h_{i},\overline{h}_{i}))\le I(Y;h_{i},\overline{h}_{i})=I(c^*(h_{i})+n;h_{i},\overline{h}_{i})=I(c^*(h_{i});h_{i})=I(Y;h_{i})
\]
Due to $h'=argmax_hI(Y;h)$ and $I(Y;h')\le I(Y;h_{i})$, we have $h'=h_{i}=argmax_hI(Y;h)$
}
Next, we prove that for $(h_{i},Y)$, if $h_{i}=argmax_hI(Y;h)$ is satisfied, then $Y=c^*(h_{i})+n$ will also be satisfied. We will prove this by contradiction. Suppose $Y\neq c^*(h_{i})+n$ and $Y=c^*(h')+n$ hold, where $h'\neq h_{i}$. Then, based on sufficiency, we have $I(c^*(h');h_{i})\le I(c^*(h');h')$. Consequently, $h'=argmax_hI(Y;h)=h_{i}$, which contradicts the given conditions. Therefore, $Y=c^*(h_{i})+n$.

\textbf{Lemma} \ref{lemma1} has been proven.

For \textbf{Lemma} \ref{lemma1}, we know that 1) the node representation $h_{v_{i}}$ (given by the GNN encoder $h_{v_{i}}=\Phi(G_v)$) satisfies the invariant condition, i.e., $P(y|h_{v_{i}},e_v)=P(y|h_{v_{i}})$ if and only if $I(y;e_v|h_{v_{i}})=0$, and 2) the node representation $h_{v_{i}}$ satisfies the necessary and sufficient condition, i.e., $Y=c^*(h_{v_{i}})+n$ if and only if $h_{v_{i}}=argmax_{h_v}I(y;h_v)$. 

We denote the GNN encoder that satisfies condition \ref{condition1} (Invariance condition) and condition \ref{condition3} (Sufficiency condition) as $\Phi^*$, and we represent the corresponding predictor model as $f^*(\cdot)=c^*(\Phi^*(\cdot))$. Let $z$ be an instance of $h_{v_{i}}$. According to assumption \ref{assumption5}, we know that there exists a random variable $\overline{z}$ such that $G_v=m(z,\overline{z})$ and $P(y|\overline{z},e_v)$ can vary arbitrarily across different environments. Based on this, for any given distribution $P_e(y,z,\overline{z})$ over environment $e$, there exists an environment $e'$ that satisfies the distribution $P_{e'}(y,z,\overline{z})$ such that
{
\abovedisplayskip=6pt 
\belowdisplayskip=6pt 
\begin{equation}
\begin{aligned}
\label{equation15}
&P_{e'}(y,z,\overline{z})=P_e(y,z)P_{e'}(\overline{z})
] 
\end{aligned}
\end{equation}
}
Then, following the reasoning line of Theorem 2 in \textcite{Wu02}, we can prove that for any function $f=c\circ \Phi$ and environment $e$, there exists an environment $e'$ such that
{
\abovedisplayskip=6pt 
\belowdisplayskip=6pt 
\begin{equation*} 
\begin{aligned}
\scalebox{0.8}{$ 
\frac{1}{|V|} \sum_{v\in V}^{}\mathbb{E}_{G_v\sim p_{e'}(\boldsymbol{G_v})}  
\mathbb{E}_{y_v\sim p_{e'}(\boldsymbol{y}|\boldsymbol{G_v}=G_v)}[l(f(G_v),y_v)]
\ge
\frac{1}{|V|} \sum_{v\in V}^{}\mathbb{E}_{G_v\sim p_{e}(\boldsymbol{G_v})}  
\mathbb{E}_{y_v\sim p_{e}(\boldsymbol{y}|\boldsymbol{G_v}=G_v)}[l(f^*(G_v),y_v)]
$}
\end{aligned}
\end{equation*}
}
Specifically, we have
{
\abovedisplayskip=6pt 
\belowdisplayskip=6pt 
\[
\begin{aligned}
& \scalebox{0.8}{$ \frac{1}{|V|} \sum_{v\in V}^{}\mathbb{E}_{G_v\sim p_{e'}(\boldsymbol{G_v})}  
\mathbb{E}_{(y_v,z_v,\overline{z}_v )\sim p_{e'}(\boldsymbol{y,z,\overline{z}}|\boldsymbol{G_v}=G_v)}
[l(c(z_v,\overline{z}_v),y_v)]$}\\&
\scalebox{0.8}{$ 
=
\frac{1}{|V'|} \sum_{v'\in V'}^{}\mathbb{E}_{G'_v\sim p_{e'}(\boldsymbol{G_v})}  
\mathbb{E}_{(\overline{z}_{v'} )\sim p_{e'}(\boldsymbol{\overline{z}}|\boldsymbol{G_v}=G_v)}[
\frac{1}{|V|} \sum_{v\in V}\mathbb{E}_{G_v\sim p_{e}(\boldsymbol{G_v})}  
\mathbb{E}_{(y_v,z_v)\sim p_{e}(\boldsymbol{y,z}|\boldsymbol{G_v}=G_v)}
[l(c(z_v,\overline{z}_{v'}),y_v)] $}\\&
\scalebox{0.8}{$\ge
\frac{1}{|V'|} \sum_{v'\in V'}^{}\mathbb{E}_{G'_v\sim p_{e'}(\boldsymbol{G_v})}  
\mathbb{E}_{(\overline{z}_{v'} )\sim p_{e'}(\boldsymbol{\overline{z}}|\boldsymbol{G_v}=G_v)}[
\frac{1}{|V|} \sum_{v\in V}\mathbb{E}_{G_v\sim p_{e}(\boldsymbol{G_v})}  
\mathbb{E}_{(y_v,z_v)\sim p_{e}(\boldsymbol{y,z}|\boldsymbol{G_v}=G_v)}
[l(c^*(z_v,\overline{z}_{v'}),y_v)]$}\\&
\scalebox{0.8}{$=
\frac{1}{|V'|} \sum_{v'\in V'}^{}\mathbb{E}_{G'_v\sim p_{e'}(\boldsymbol{G_v})}  
\mathbb{E}_{(\overline{z}_{v'} )\sim p_{e'}(\boldsymbol{\overline{z}}|\boldsymbol{G_v}=G_v)}[
\frac{1}{|V|} \sum_{v\in V}\mathbb{E}_{G_v\sim p_{e}(\boldsymbol{G_v})}  
\mathbb{E}_{(y_v,z_v)\sim p_{e}(\boldsymbol{y,z}|\boldsymbol{G_v}=G_v)}
[l(c^*(z_v),y_v)]$}\\&
\scalebox{0.8}{$=
\frac{1}{|V|} \sum_{v\in V}^{}\mathbb{E}_{G_v\sim p_{e}(\boldsymbol{G_v})}  
\mathbb{E}_{y_v\sim p_{e}(\boldsymbol{y}|\boldsymbol{G_v}=G_v)}[l(f^*(G_v),y_v)]
]$}
\end{aligned}
\]
}
The first equality is given by Eqn. (\ref{equation15}), and the second/third steps are due to the sufficiency condition (condition \ref{condition3}) of $\Phi^*$. Therefore, a function $f^*$ that satisfies both condition \ref{condition1} and condition \ref{condition3} will have optimal performance across all environments.

\subsection{NV-REx}
\label{C.3}
Here, we will demonstrate that our invariant regularizer NV-REx, provides a tighter upper bound compared to V-REx.

$Proof$. First, from a node-level perspective, let's redefine the KL distance as:
{
\abovedisplayskip=6pt 
\belowdisplayskip=6pt 
\begin{equation}
\begin{aligned}
\label{equation16}
&D_{KL}(p_e(\boldsymbol{y|G_v})||q(\boldsymbol{y|G_v})):=\frac{1}{|V|} \sum_{v\in V}^{}\mathbb{E}_{G_v\sim p_e(\boldsymbol{G_v})}  
\mathbb{E}_{y_v\sim p_e(\boldsymbol{y}|\boldsymbol{G_v}=G_v)}[log\frac{p_e(\boldsymbol{y}=y_v|\boldsymbol{G_v}=G_v)}{q(\boldsymbol{y}=y_v|\boldsymbol{G_v}=G_v)} ]
\end{aligned}
\end{equation}
}
We utilize $D_{KL}(q(\boldsymbol{y|z})||\mathbb{E}_{\boldsymbol{e}}[ q(\boldsymbol{y|z})])$ to measure the generalization error,
{
\abovedisplayskip=6pt 
\belowdisplayskip=6pt 
\[
\begin{aligned}
&D_{KL}(q(\boldsymbol{y|z})||\mathbb{E}_{\boldsymbol{e}}[ q(\boldsymbol{y|z})])\\&
\scalebox{0.8}{$ \frac{1}{|V|} \sum_{v\in V}^{}\mathbb{E}_{\boldsymbol{e}} \mathbb{E}_{G\sim p_e(\boldsymbol{G})} 
\mathbb{E}_{y_v\sim p_e(\boldsymbol{y|G_v}=G_v)} 
\mathbb{E}_{z_v\sim p_e(\boldsymbol{z|G_v}=G_v)}
[log\frac{q(\boldsymbol{y}=y_v|\boldsymbol{z}=z_v)}{\mathbb{E}_{\boldsymbol{e}}
[q(\boldsymbol{y}=y_v|\boldsymbol{z}=z_v)]} ]$}\\&
\scalebox{0.8}{$=
\frac{1}{|V|} \sum_{v\in V}^{}\mathbb{E}_{\boldsymbol{e}} \mathbb{E}_{G\sim p_e(\boldsymbol{G})} 
\mathbb{E}_{y_v\sim p_e(\boldsymbol{y|G_v}=G_v)} 
\mathbb{E}_{z_v\sim p_e(\boldsymbol{z|G_v}=G_v)}
[log\ q(\boldsymbol{y}=y_v|\boldsymbol{z}=z_v)-log\mathbb{E}_{\boldsymbol{e}}
\left[q(\boldsymbol{y}=y_v|\boldsymbol{z}=z_v)\right] ]$}\\&
\scalebox{0.8}{$\le
\frac{1}{|V|} \sum_{v\in V}^{}\mathbb{E}_{\boldsymbol{e}} \mathbb{E}_{G\sim p_e(\boldsymbol{G})} 
\mathbb{E}_{y_v\sim p_e(\boldsymbol{y|G_v}=G_v)} 
\mathbb{E}_{z_v\sim p_e(\boldsymbol{z|G_v}=G_v)}
[log\ q(\boldsymbol{y}=y_v|\boldsymbol{z}=z_v)-\mathbb{E}_{\boldsymbol{e}}
\left[log\ q(\boldsymbol{y}=y_v|\boldsymbol{z}=z_v)\right] ]$}
\end{aligned}
\]
}
Using Jensen Inequality, we can also obtain an upper bound for $D_{KL}(q(\boldsymbol{y|z})||\mathbb{E}_{\boldsymbol{e}}[ q(\boldsymbol{y|z})])$ as
{
\abovedisplayskip=6pt 
\belowdisplayskip=6pt 
\begin{equation}
\begin{aligned}
\label{equation17}
& {\textstyle \sum_{v\in V}^{}} \mathbb{E}_{\boldsymbol{e}}|l(f(G_v),y_v)
-\mathbb{E}_{\boldsymbol{e}}[l(f(G_v),y_v)]| 
\end{aligned}
\end{equation}
}
We optimize its squared form, denoted as \(\mathbb{V}_{node-wise} = {\textstyle \sum_{v\in V}^{}}\mathbb{E}_{\boldsymbol{e}}[l(f(G_v),y_v)-\mathbb{E}_{\boldsymbol{e}}[l(f(G_v),y_v)] ]^2  \). 
Similarly, utilizing Jensen Inequality, we have:
{
\abovedisplayskip=6pt 
\belowdisplayskip=6pt 
\[
\begin{aligned}
& \mathbb{V}_{node-wise} =\\&
\scalebox{0.8}{$\frac{1}{|V|} \sum_{v\in V}^{}\mathbb{E}_{\boldsymbol{e}} \mathbb{E}_{G\sim p_e(\boldsymbol{G})} 
\mathbb{E}_{y_v\sim p_e(\boldsymbol{y|G_v}=G_v)} 
\mathbb{E}_{z_v\sim p_e(\boldsymbol{z|G_v}=G_v)}
[log\ q(\boldsymbol{y}=y_v|\boldsymbol{z}=z_v)-\mathbb{E}_{\boldsymbol{e}}
\left[log\ q(\boldsymbol{y}=y_v|\boldsymbol{z}=z_v)\right] ]$}\\&
\scalebox{0.8}{$\le
\mathbb{E}_{\boldsymbol{e}} \frac{1}{|V|} \sum_{v\in V}^{}\mathbb{E}_{G\sim p_e(\boldsymbol{G})} 
\mathbb{E}_{y_v\sim p_e(\boldsymbol{y|G_v}=G_v)} 
\mathbb{E}_{z_v\sim p_e(\boldsymbol{z|G_v}=G_v)}
[log\ q(\boldsymbol{y}=y_v|\boldsymbol{z}=z_v)-\mathbb{E}_{\boldsymbol{e}}
\left[log\ q(\boldsymbol{y}=y_v|\boldsymbol{z}=z_v)\right] ] $}\\&
=\mathbb{V}_{graph-wise} \\&
\end{aligned}
\]
}
Therefore, optimizing NV-REx ($\mathbb{V}_{node-wise}$) compared to V-REx ($\mathbb{V}_{graph-wise}$) allows the generalization error to converge to a tighter upper bound.

\section{Ablation Experiments}
\label{D}
\subsection{Supplementary Table}
\label{D.1}
\begin{table}[h]
\caption{Efficiency Comparison. IENE-re achieves superior OOD generalization performance and demonstrates better time and space efficiency compared to EERM. Experiments were conducted on Nvidia A40 (48G).}
\label{table3}
\vskip 0.0in
\begin{center}
\begin{small}
\begin{tabular}{c|cccc|cccc}
\hline
\ & \multicolumn{4}{|c}{Running Time (s/100 epochs)} & \multicolumn{4}{|c}{Total GPU Memory(GB)} \\
\hline
\  & Cora& Amz-Photo& Elliptic& OGB-Arxiv& Cora& Amz-Photo& Elliptic& OGB-Arxiv\\
\hline
TIVA & 7& 13& 31&47&2.23&3.71&4.02&5.49\\
IENE-r & 9& 21& 45&62&2.41&3.97&4.76&8.69\\
EERM & 30& 174& 280&440&4.19&18.79&19.78&33.64\\
IENE-e & 13& 62& 130&311&3.02&8.27&10.81&25.24\\
IENE-re & 14& 66&137&328&3.03&8.48&10.99&25.36\\
\hline
\end{tabular}
\end{small}
\end{center}
\vskip -0.1in
\end{table}
\subsection{Selection of Auxiliary Information}
\label{D.2}
To assess the influence of auxiliary information on performance, we performed ablation studies on the synthetic dataset described in Section \ref{section4}. We specifically employed environmental representation $h_{v_{e}}$, irrelevant features ${X_{ir}}$, and random noise $n$ as auxiliary information for environmental partitioning. The informative content of each auxiliary information was evaluated by measuring the average accuracy. For identifying irrelevant features ${X_{ir}}$, we adopted the methodology proposed by TIVA \cite{Tan}, while standard Gaussian noise was utilized for the random noise $n$. The experimental settings were consistent with those described in Section \ref{section4.2}. The final experimental results are presented in Table \ref{table4}.
\begin{table}[h]
\caption{The ablation study reults on different selection of auxiliary information.}
\label{table4}
\vskip 0.0in
\begin{center}
\begin{small}
\begin{tabular}{c|ccc}
\hline
\ & $h_{v_{e}}$& $X_{ir}$& $n$\\
\hline
Cora & \textbf{95.58$\pm$ 2.07}& 94.33$\pm$2.18 &91.41$\pm$3.59\\
Amz-Photo & \textbf{96.40 $\pm$1.80}&95.45 $\pm$2.60&91.78 $\pm$1.73\\
\hline
\end{tabular}
\end{small}
\end{center}
\vskip -0.1in
\end{table}

Based on the results of the ablation study, we observed that using $h_{v_{e}}$ for environmental recognition yielded the best performance, while using $X_{ir}$ resulted in suboptimal performance. This can be attributed to the fact that $h_{v_{e}}$ not only learns from $X_{ir}$ but also incorporates rich spurious feature information that is causally related to the environment. These spurious features play a direct and key role in environmental partitioning. Additionally, our ablation study indicates that the model cannot utilize random noise as auxiliary information since it does not contain any information about $X$, thus it is unable to assist in distinguishing the spurious features. It is worth noting that when no useful component is found in the auxiliary information, the model would degrade to ERM (Empirical Risk Minimization).
\subsection{V-REx and NV-REx}
\label{D.3}
According to Appendix \ref{C.3}, we have theoretically proven that the node-level variance (NV-REx) possesses a tighter upper bound compared to the graph-level variance (V-REx). Here, we further validate their performance as invariant penalties. We employ GCN as the backbone network and train the models using the IENE-re method on six datasets. The experimental settings are identical, except for the invariant penalty. The results are presented in Table \ref{table5}.
\begin{table}[h]
\caption{The ablation study reults on different selection of variance penalty.}
\label{table5}
\vskip 0.0in
\begin{center}
\begin{small}
\begin{tabular}{c|cc}
\hline
\ & V-REx& NV-REx\\
\hline
Cora & 94.71$\pm$ 2.29& \textbf{95.58$\pm$2.07} \\
Amz-Photo & 95.48 $\pm$1.32&\textbf{96.40 $\pm$1.80} \\
Twitch-E & 59.57 $\pm$2.51&\textbf{61.87 $\pm$1.43} \\
FB-100 & 52.46 $\pm$1.69&\textbf{53.73 $\pm$2.65} \\
Elliptic & 66.85 $\pm$1.54&\textbf{67.90 $\pm$3.31} \\
OGB-Arxiv & 44.19 $\pm$0.90&\textbf{44.85 $\pm$0.33} \\
\hline
\end{tabular}
\end{small}
\end{center}
\vskip -0.1in
\end{table}

As shown in Table \ref{table5}, NV-REx demonstrates better performance on all datasets. This indicates that in the task of node classification on graphs under out-of-distribution (OOD) settings, NV-REx achieves better performance as it can identify a greater amount of environmental variations.
\subsection{Method of Environment Extrapolation}
\label{D.4}
We explored different graph augmentation methods and validated the effectiveness of using IENE-re for extrapolating beyond known environments. It is important to note that EERM and FLOOD generate augmented views by randomly adding or removing edges. We will compare the following methods for generating environment views: a method based on random edge dropout \cite{Waniek}, a method based on gradient perturbation of edges \cite{Xu02}, and EERM \cite{Wu02}. Each method has the same edge adjustment budget. It is worth emphasizing that these methods only differ in their view generation approaches, while the training framework remains the same as IENE-re. The results are presented in Table \ref{table6}. 
\begin{table}[h]
\caption{The ablation study results on different methods of environment extrapolation.}
\label{table6}
\vskip 0.0in
\begin{center}
\begin{small}
\begin{tabular}{c|cc}
\hline
\ & Cora& Amaz-Photo\\
\hline
Random & 91.43$\pm$ 3.96& 93.17$\pm$1.63 \\
Grad-based & 93.49 $\pm$2.50&95.38 $\pm$1.67\\
EERM & 94.77 $\pm$1.85&95.14 $\pm$1.46\\
IENE-re & \textbf{95.58 $\pm$2.07}& \textbf{96.40 $\pm$1.80} \\
\hline
\end{tabular}
\end{small}
\end{center}
\vskip -0.1in
\end{table}

Several general methods for augmenting graph data have been proposed, but they can not guarantee the quality of the generated views. Specifically, from an invariance perspective, it is desirable for the generated views to maximize the differences in spurious features, thereby fully satisfying Condition \ref{condition2} (Non-Invariance Discrimination Condition). Existing data augmentation methods do not meet these requirements and are not well-suited for our research. In contrast, IENE-re excels by generating augmented views that better adhere to these conditions, resulting in superior performance.
\subsection{Different Frameworks of IENE}
\label{D.5}
We compared the performance of IENE-r, IENE-e, and IENE-re on six datasets. The results of the experimental evaluation are shown in Table \ref{table7}.
\begin{table}[h]
\caption{The ablation study reults on different frameworks of IENE.}
\label{table7}
\vskip 0.0in
\begin{center}
\begin{small}
\resizebox{1.0\columnwidth}{!}{

\begin{tabular}{cccccccc}
\hline
Backbone&  Method& Cora& Amz-Photo& Twitch-E& FB-100& Elliptic& OGB-Arxiv\\ 
\hline
\multirow{3}{*}{GCN}
& IENE-r& 94.76$\pm$ 2.31& 95.49$\pm$1.64& 60.49$\pm$ 1.51& 51.98$\pm$1.57& 66.18$\pm$0.85& 43.37$\pm$0.50\\
& IENE-e& 95.13$\pm$ 1.52& 96.24$\pm$1.15& 61.10$\pm$ 0.88& 51.42$\pm$2.34& 66.79$\pm$1.64& 43.98$\pm$1.21\\
& IENE-re& \textbf{95.58$\pm$ 2.07}& \textbf{96.40$\pm$1.80}& \textbf{61.87$\pm$ 1.43}& \textbf{53.73$\pm$2.65}& \textbf{67.90$\pm$3.31}& \textbf{44.85$\pm$0.31}\\
\hline
\multirow{3}{*}{SAGE}
& IENE-r& 98.31$\pm$ 0.90& 95.21$\pm$1.44& 64.31$\pm$ 0.82& OOM& 65.80$\pm$4.08& 40.67$\pm$1.90\\
& IENE-e& 98.68$\pm$ 0.48& 96.48$\pm$0.67& 64.96$\pm$ 0.60& OOM& 66.72$\pm$4.81& 41.58$\pm$1.63\\
& IENE-re& \textbf{99.27$\pm$ 0.55}& \textbf{96.87$\pm$1.65}& \textbf{66.18$\pm$ 0.78}& OOM& \textbf{68.38$\pm$3.78}& \textbf{41.82$\pm$1.84}\\
\hline
\multirow{3}{*}{GAT}
& IENE-r& 96.86$\pm$ 1.32& 95.97$\pm$0.40& 62.19$\pm$ 1.79& 53.61$\pm$0.73& 65.65$\pm$3.42& 44.37$\pm$1.49\\
& IENE-e& 97.92$\pm$ 0.47& 95.48$\pm$1.51& 62.46$\pm$ 1.03& 53.98$\pm$1.59& 64.73$\pm$2.76& 44.17$\pm$2.61\\
& IENE-re& \textbf{99.00$\pm$ 0.60}& \textbf{96.85$\pm$1.76}& \textbf{62.55$\pm$ 1.42}& \textbf{54.72$\pm$ 2.05}& \textbf{66.56$\pm$3.95}& \textbf{45.70$\pm$0.81}\\
\hline
\multirow{3}{*}{GPR}
& IENE-r& 91.97$\pm$ 3.69& 91.75$\pm$1.58& 63.44$\pm$ 0.14& 54.55$\pm$1.70& 62.73$\pm$1.69& 47.24$\pm$0.99\\
& IENE-e& 93.39$\pm$2.82& 92.36$\pm$3.16&62.87$\pm$0.37& 54.91$\pm$1.85& 62.89$\pm$0.94& 47.03$\pm$1.28\\
& IENE-re& \textbf{94.36$\pm$ 3.34}& \textbf{92.93$\pm$2.76}& \textbf{64.68$\pm$ 0.72}& \textbf{55.75$\pm$ 1.28}& \textbf{63.57$\pm$1.65}& \textbf{47.53$\pm$1.20}\\
\hline
\end{tabular}
}
\end{small}
\end{center}
\vskip -0.1in
\end{table}

The table clearly indicates that the IENE-re outperforms others on all datasets and backbones. This superiority can be attributed to the fact that using IENE-r or IENE-e independently has certain limitations that lead to a decrease in performance. Specifically, IENE-r's identified invariant features may not exhibit sufficient tolerance to structural variations. On the other hand, IENE-e struggles to allocate weights to the classification loss for different environments as flexibly and precisely as IENE-r does. Consequently, it fails to fully maximize the invariant penalty and disregards certain invariant features. In contrast, IENE-re effectively combines the strengths of both methods, resulting in more efficient and robust invariant learning.
\subsection{Hyperparameter Analysis}
\label{D.6}
We primarily analyze three key hyperparameters of the model, namely, $K$, $\lambda$, and $\beta$.

\subsubsection{Influence of $K$}
We conducted an ablation analysis on the synthetic dataset to explore the impact of the value of $K$ on our proposed algorithm. It is important to note that when $K=1$, the model degenerates into ERM (Empirical Risk Minimization). Here, we report the average accuracy for each test. The computational results are presented in Table \ref{table8}.
\begin{table}[h]
\caption{The ablation study results on different value of $K$ on synthetic dataset.}
\label{table8}
\vskip 0.0in
\begin{center}
\begin{small}
\begin{tabular}{cc}
\hline
\ & IENE-re\\
\hline
$K=2$ & 95.13 \\
$K=3$ & \textbf{95.58}\\
$K=4$ & 94.79\\
$K=5$ &  94.90\\
$ERM$ &  92.29\\
\hline
\end{tabular}
\end{small}
\end{center}
\vskip -0.1in
\end{table}

The results of the ablation experiments demonstrate that our algorithm shows relative flexibility in the value of $K$, which is similar to the observations in TIVA \cite{Tan} and ZIN \cite{Lin01}.

\subsubsection{Influence of $\lambda$}
\begin{figure}[ht]
\vskip 0.0in
\begin{center}
\centerline{\includegraphics[scale=0.25]{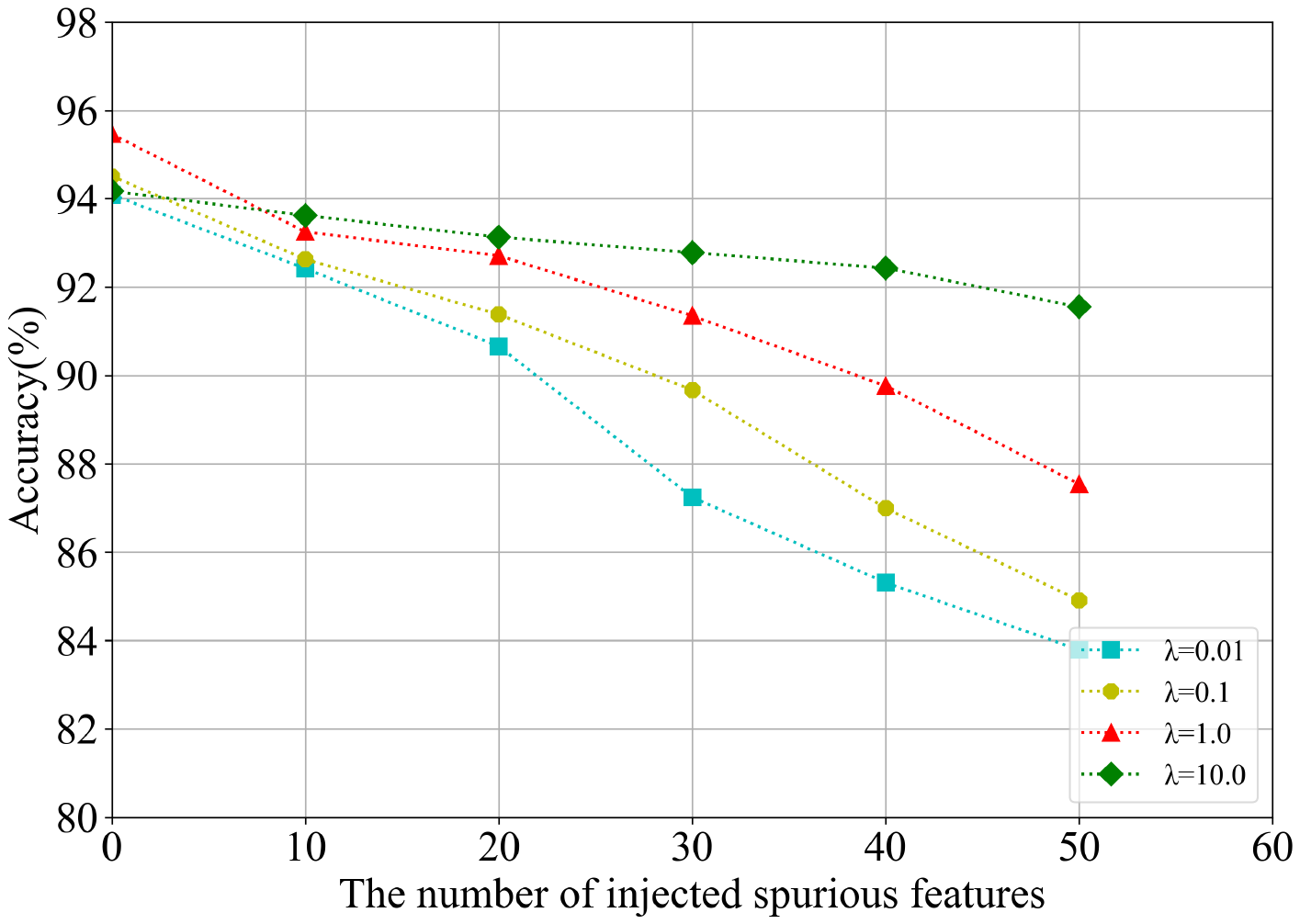}}
\caption{Ablation experiments on the value of parameter $\lambda$.}
\label{figure3}
\end{center}
\vskip -0.2in
\end{figure}

Based on the discussion in Appendix \ref{B5}, our model allowed us to control the degree to which the model learns confounding invariances, as shown by the $X_2$ in Figure \ref{figure1}(c), by adjusting the coefficient of the invariant penalty term. To validate the aforementioned observation, we performed experimental adjustments to the parameter $\lambda$ and employed the offset technique proposed by \textcite{Wu02}. This technique can generate additional synthetic features that possess similar characteristics to $X_4$ in Figure \ref{figure1}(c). The results of these experiments are presented in Figure \ref{figure3}. 

Based on Figure \ref{figure3}, we can draw the following conclusions: when no additional spurious features are added, the model's performance improves as $\lambda$ increases within a certain range ($\lambda <1.0$). Besides, for sufficiently large values of $\lambda$ (e.g., $\lambda \ge 1.0$), larger values of $\lambda$ (such as $\lambda =10.0$) help force the model to learn knowledge from features that are completely unaffected by the environment, as represented by $X_1$ in Figure \ref{figure1}(c). This comes at the cost of sacrificing a certain level of accuracy since some features that are causally related to the prediction $Y$, such as $X_2$, are ignored, which to some extent violates the sufficiency condition (Condition \ref{condition3}). However, the model exhibits inclusiveness towards additionally added confusable spurious features. On the other hand, smaller values of $\lambda$ (e.g., $\lambda =1.0$) allow the model to learn knowledge from $X_2$, resulting in better initial performance. However, as the number of confusable spurious features similar to $X_2$ (such as $X_4$) increases, the OOD (out-of-distribution) performance deteriorates at a relatively faster rate due to the difficulty of distinguishing between $X_2$ and $X_4$.
\subsubsection{Influence of $\beta$}
We analyzed the impact of hyperparameter $\beta$ on performance, as shown in Table \ref{table9}. The experiments were conducted on the synthetic dataset Cora. Here, we report the average accuracy for each test. Additionally, to validate the analysis in Section \ref{B2}, we randomly perturbed 5\% of the edges in the test graphs based on DICE \cite{Waniek} and evaluated the model's sensitivity to structural changes. 
\begin{table}[h]
\caption{The ablation study reults on different value of parameter $\beta$.}
\label{table9}
\vskip 0.0in
\begin{center}
\begin{small}
\begin{tabular}{ccccccc}
\hline
\ IENE-re & $\beta=0$&$\beta=0.01$&$\beta=0.1$&$\beta=1.0$&$\beta=5.0$&$\beta=10.0$\\
\hline
Original test data & 93.99& 94.62& 95.26& \textbf{95.58}& 95.17& 94.21\\
+Random noise & 91.73& 92.51& 93.28& \textbf{93.93}& 93.67& 92.78\\
Decrease & 2.26& 2.11& 1.98& 1.65& 1.50& \textbf{1.43}\\
\hline
\end{tabular}
\end{small}
\end{center}
\vskip -0.1in
\end{table}

The results indicate that choosing the appropriate value for parameter $\beta$ is a determining factor for the model's performance. As $\beta$ increases within a certain range, the OOD (out-of-distribution) performance improves. Additionally, larger values of $\beta$ tend to enable the model to learn invariant features that can adapt to variations of structural environment.

\subsection{More Comparisons}
\label{D.7}
Due to space limitations in the main text, we will present a comparison of some methods here. As shown in Table \ref{table10}, our method still maintains superior performance. 

\begin{table}[h]
\caption{Comparison between IENE and other methods using GCN as Backbone.}
\label{table10}
\vskip 0.0in
\begin{center}
\begin{small}
\begin{tabular}{c|c|c|c|c|c|c}
\hline
\ Method & Cora& Amaz-Photo & Twitch-E & FB-100& Elliptic& OGB-Arxiv\\
\hline
SRGNN & 92.43$\pm$ 2.48& 93.85$\pm$1.31& 56.04$\pm$1.63& 52.91$\pm$1.60& 64.76$\pm$3.59& 44.24$\pm$1.78 \\
LISA & 93.85 $\pm$1.91&94.22 $\pm$1.06& 57.83$\pm$2.16& 53.17$\pm$1.39& 64.39$\pm$2.62& 43.49$\pm$0.71\\
IENE & \textbf{95.58 $\pm$2.07}& \textbf{96.40 $\pm$1.80} & \textbf{61.87 $\pm$1.43}& \textbf{53.73 $\pm$2.65}& \textbf{67.90 $\pm$3.31}& \textbf{44.85 $\pm$0.31}\\
\hline
\end{tabular}
\end{small}
\end{center}
\vskip -0.1in
\end{table}

\subsection{Performance on In-distribution Data and Different Subsets}
\label{D.8}
To verify the stability of our model on different data, we have supplemented the experimental results with in-distribution data on six datasets and various test subsets on Twitch and Arxiv, as shown in Table \ref{table11} to \ref{table13}.  

\begin{table}[h]
\caption{The performance of IENE on In-distribution dataset.}
\label{table11}
\vskip 0.0in
\begin{center}
\begin{small}
\begin{tabular}{c|c|c|c|c|c|c}
\hline
\ & Cora& Amaz-Photo& Twitch-E & FB-100& Elliptic& OGB-Arxiv\\
\hline
ERM & 94.80$\pm$ 0.25& 95.49$\pm$0.21& \textbf{71.41$\pm$0.03}& 59.31$\pm$0.32& 74.87$\pm$0.12& 47.21$\pm$0.45 \\
IENE & \textbf{97.33 $\pm$0.27}&\textbf{96.71 $\pm$0.38}& 71.27$\pm$0.08& \textbf{59.52$\pm$0.40}& \textbf{75.18$\pm$0.48}& \textbf{48.73$\pm$0.30} \\

\hline
\end{tabular}
\end{small}
\end{center}
\vskip -0.1in
\end{table}

\begin{table}[h]
\caption{The performance of IENE on different subgraphs of Twitch.}
\label{table12}
\vskip 0.0in
\begin{center}
\begin{small}
\begin{tabular}{c|c|c|c|c|c}
\hline
\ & ES& FR& PTBR & RU& TW \\
\hline
ERM & 60.73$\pm$ 2.75& 56.41$\pm$2.14& 59.39$\pm$1.80& 45.77$\pm$1.55 & 47.90$\pm$1.76 \\
IENE & \textbf{63.74 $\pm$1.43}&\textbf{61.25 $\pm$1.05}& \textbf{64.85$\pm$1.11}& \textbf{56.26$\pm$0.52}& \textbf{57.59$\pm$0.83} \\

\hline
\end{tabular}
\end{small}
\end{center}
\vskip -0.1in
\end{table}

\begin{table}[h]
\caption{The performance of IENE on different publication years of Arxiv.}
\label{table13}
\vskip 0.0in
\begin{center}
\begin{small}
\begin{tabular}{c|c|c|c}
\hline
\ & 2014-2016& 2016-2018& 2018-2020 \\
\hline
ERM & 42.83$\pm$ 0.93& 41.58$\pm$1.67& 40.36$\pm$0.79 \\
IENE & \textbf{46.39 $\pm$0.44}&\textbf{44.70 $\pm$0.92}& \textbf{43.46$\pm$0.53} \\

\hline
\end{tabular}
\end{small}
\end{center}
\vskip 0.0in
\end{table}

\section{Hyperparameter Settings and Experimental Details}
\label{E}

\begin{table}[h]
\caption{Hyperparameter settings for the experiments.}
\label{table14}
\vskip -0.0in
\begin{center}
\begin{small}
\resizebox{1.0\columnwidth}{!}{
\begin{tabular}{cccccccc}
\hline
\ Dataset & $Learning\ rate$&$K$&$\lambda$&$\beta$& $\eta$& $Learning\ rate for\ u, d, w$& $num\_sample$\\
\hline
Cora & [0.005,0.01]& [2,3,4]& 4.0& 1.0& 1.0& [0.005,0.01,0.05]& [1,3,5,7]\\
Amz-Photo & [0.01,0.05]& [3,4]& 4.0& 1.0& 1.0& [0.001,0.1,0.5]& [1,3,5,7]\\
Twitch-E & [0.01,0.05]& [2,3,4]& 4.0& 3.0& 1.0& [0.05,0.1]& [1,3,5,7]\\
FB-100 & [0.005,0.01]& [2,3,4]& 4.0& 1.0& 1.0&  [0.05,0.1]& [1,3,5,7]\\
Elliptic & [0.005,0.01]& [2,3,4]& 1.0& 0.5& 1.0&  [0.01,0.05,0.1]& [1,3,5,7]\\
OGB-Arxiv & [0.001,0.01]& [2,3,4]& 1.0& 1.0& 1.0&  [0.05,0.1]& [1,3,5,7]\\
\hline
\end{tabular}
}
\end{small}
\end{center}
\vskip -0.1in
\end{table}
In this section, we will present the experimental details of Section \ref{section4}. All experiments were conducted on an Nvidia A40 (48G) GPU, and the hyperparameters used for the methods proposed in Section \ref{section4} are shown in Table \ref{table14}. The encoders and decoders used in all experiments are GNNs, and the classifier is a 2-layer MLP with 32 neurons. Here, we use the ReLU activation function \cite{Li04} in all the networks. The output layers of the encoders $u(\cdot)$ and $\Phi(\cdot)$ are activated using the tanh function \cite{Xiao}. Hyperparameters in different network may be different, and we fine-tune these hyperparameters through a grid search method across 10 trials. 

We primarily utilize the Adam optimizer for parameter updates \cite{Kingma}, except for $u(\cdot)$, $\Phi(\cdot)$, $d(\cdot)$ and $w(\cdot)$ which use the SGD optimizer \cite{Bottou} to enhance the robustness of adversarial training.

The complete and reproducible code for our work has been anonymously submitted to:
https://anonymous.4open.science/r/IENE-code

\section{\textbf{\textbf{Related Work}}}
\label{F}
\subsection{\textbf{\textbf{Invariant Learning}}}
\label{F.1}
Invariant learning is a technique aimed at discovering invariant features to achieve out of distribution (OOD) generalization. Invariant features, denoted as \(X_{i}\), exhibit stable correlations with the target \textit{Y} when distribution shifts, while spurious features \(X_{s}\) are unstable. In general, models relying on invariant features maintain robust on OOD data due to stable correlations when violating the independent and identically distributed (IID) assumption. Conversely, spurious features \(X_{s}\) can lead to significant performance degradation \cite{Arjovsky}. \textcite{Peters}, starting from causal theory, first investigated the fact that invariance can infer causal structures under necessary conditions, proposing Invariant Causal Prediction (ICP). Invariant Risk Minimization (IRM) is introduced to learn the optimal invariant associations across different environment partitions provided in the training data \cite{Arjovsky}. To further explore the capabilities of IRM, some literature combines multi-objective optimization with game theory \cite{Ahuja,Chen02} or conducts invariant representation learning through adversarial training \cite{Chang,Xu01}. REx \cite{Krueger} encourages reducing training risk while increasing similarity in training risk under the variance risk extrapolation paradigm. GroupDRO \cite{Sagawa} has a similar problem setting to IRM, which obtains a robust model by reducing the worst-case training loss on a predefined set of groups.

However, most existing methods, including the aforementioned ones, greatly rely on additional environment labels explicitly provided in the training dataset. Such annotations for nodes on graph data are often unavailable and costly to collect, making these invariant learning methods impractical for real-world data \cite{Creager,Liu01}. It has led to multiple efforts in performing invariant learning without the need for explicitly provided environment partitions. \textcite{Sohoni} utilize clustering techniques to partition the training data into different groups, then incorporate the labels of these groups as noise supervision into a distributionally robust optimization objective. EIIL \cite{Creager} addresses the issue through a two-stage training involving biased model environment reasoning and sequence invariance learning for environment reasoning. Heterogeneous Risk Minimization (HRM) \cite{Liu01} jointly learns latent heterogeneity and invariance by using an identification module and an invariance predictor, respectively. ZIN \cite{Lin01} demonstrates that additional auxiliary variables provided by metadata of the dataset can be used for environment inference and subsequent invariant learning. TIVA \cite{Tan} attempts to automatically learn environment partitions using target-independent variable \(X_{z} \) (\(X_{z} \bot Y\)). However, on one hand, the performance of environment partitioning is limited due to the insufficient discovery of diverse environment factors; on the other hand, the assumption that \(X_{z} \) can be used to partition the environment remains challenging to verify because \(X_{z} \) may not have explicit physical meanings in many datasets.

In this work, we learn both environmental representations and invariant representations in a disentangled information bottleneck framework. We use environmental representation as auxiliary information for environmental reasoning and invariant representation as input for downstream tasks. Compared to the aforementioned methods, our approach is conducive to extracting more environmental information and promoting mutual learning between environmental learning and invariant learning. 

\subsection{\textbf{\textbf{Invariant Learning on Graph}}}
\label{F.2}
Invariant learning on graphs can be more complex due to the various forms of distribution shifts, such as in features and structures, making it challenging to identify invariance. Recently, there has been widespread interest in exploring the out-of-distribution (OOD) generalization capability of GNNs. In particular, \textcite{Bevilacqua} focused on learning size-invariant representations to address distribution shifts related to graph size. DIR \cite{Wu03} created multiple randomly paired intervention graphs to achieve the goal of invariant causal representation learning across different distributions, enhancing generalization. GIL \cite{Li03}, through joint optimization of three custom modules, captured the invariant relationships between the structural information and labels of predicted graphs in mixed latent environments. CIGA \cite{Chen03} used causal models to characterize changes in distribution. MoleOOD \cite{Yang01} improved the robustness of molecular learning and could infer the environment in a data-driven manner. LISA \cite{Lisa} proposes a model-agnostic label-invariant subgraph augmentation framework to generate augmented environments with consistent predictive relationships for graph OOD generalization. These efforts have primarily centered on graph-level tasks, emphasizing the extraction of core subgraphs while generally overlooking the more challenging node-level tasks with multiple latent environments. In contrast, our work addresses the OOD problem in node-level tasks on graphs.

Some works addressing semi-supervised node classification in non-I.I.D. settings have been proposed to focus on the adaptability of GNNs under distribution shifts. For instance, SR-GNN \cite{Zhu} is designed to interpret the distribution differences between biased training data and the true inference distribution of the graph. CATs \cite{He} introduced a joint attention mechanism considering structural perturbations to enhance the model's generalization to structural biases. BA-GNN \cite{Chen01} improved the model's generalization by identifying biases and learning invariant features. DGNN \cite{Fan} introduced the idea of variable decorrelation in GNNs to mitigate the impact of label selection bias. However, designing models specifically for certain types of biases may not truly enhance generalization ability in real-world applications due to the presence of confounding biases. EERM \cite{Wu02}, learns invariant representations of nodes by training multiple context generators to maximize the risk differences from multiple virtual environments. FLOOD \cite{Liu02} proposes an invariant learning component to construct multiple environments from graph data augmentation and learns invariant representations under risk extrapolation. However, EERM and FLOOD cannot guarantee the quality of the generated views during data augmentation, making it difficult to provide reliable support for invariant learning. Because view generators may take shortcuts to achieve their objectives, such as increasing risk differences in different environments by adding varying degrees of noise or reducing the amount of information, violating the assumption of environmental heterogeneity. CIE \cite{Chen04} introduces the independence constraint to improve discriminability and stability of causal and spurious features in complex biased environments and mitigates spurious correlations by implementing the backdoor adjustment. CaNet \cite{QTWu} uses a new learning objective derived from causal inference that coordinates an environment estimator and a mixture-of-expert GNN predictor to counteract the confounding bias in training data. These two works combine the idea of causal reasoning, aiming to eliminate confounding bias in the data.

\section{\textbf{\textbf{Limitations}}}
\label{G}
The limitations of IENE have been partially discussed in sections \ref{Section3.3.2} and \ref{B5} that we discussed the scenarios where IENE can successfully identify invariant features. Here, we will discuss some other limitations. 

\textbf{Dependence on assumptions.} The success of IENE still relies on some common assumptions that are required in many invariant learning works. Therefore, in some specific cases, it may not be able to recognize all invariant features. 

\textbf{Requirements for computing resource.} Although the environmental extrapolation module can potentially enhance generalization performance by identifying more invariant features, it significantly increases the model's time and memory overhead on some datasets, as shown in Table \ref{table3}. This may not be user-friendly for resource-constrained scenarios. 

 \textbf{Scalability.} The scalability of the model on large-scale and complex graphs remains to be further validated. We believe that IENE possesses scalability since it is a relatively flexible learning framework where the backbone network can be freely switched and potentially integrated with other frameworks. Moreover, our empirical results indicate that switching the backbone network does not affect the effectiveness of the framework. Therefore, it is possible to extend IENE to large-scale graphs by employing local graph sampling strategies or sampling-based networks such as GraphSAGE as the backbone. Similarly, adopting heterogeneous graph convolutional networks as the backbone can enable extension to heterogeneous graphs. While this idea seems simple, it may be feasible, and we plan to further validate the capabilities of IENE on large-scale and heterogeneous graph datasets in the future. 

\newpage
\section*{NeurIPS Paper Checklist}

\begin{enumerate}

\item {\bf Claims}
    \item[] Question: Do the main claims made in the abstract and introduction accurately reflect the paper's contributions and scope?
    \item[] Answer: \answerYes{} 
    \item[] Justification: We confirm that the main claims presented in the abstract and introduction reflect the contribution and scope of the paper.

\item {\bf Limitations}
    \item[] Question: Does the paper discuss the limitations of the work performed by the authors?
    \item[] Answer: \answerYes{} 
    \item[] Justification: We discussed the limitations of our work in the appendix \ref{G}.

\item {\bf Theory Assumptions and Proofs}
    \item[] Question: For each theoretical result, does the paper provide the full set of assumptions and a complete (and correct) proof?
    \item[] Answer: \answerYes{} 
    \item[] Justification: We have listed the used assumptions in Appendix \ref{B6} and provided proof in Appendix \ref{C}.

    \item {\bf Experimental Result Reproducibility}
    \item[] Question: Does the paper fully disclose all the information needed to reproduce the main experimental results of the paper to the extent that it affects the main claims and/or conclusions of the paper (regardless of whether the code and data are provided or not)?
    \item[] Answer: \answerYes{} 
    \item[] Justification: We provide detailed model parameter information in Appendix \ref{E} and attach reproducible code. And the dataset used in the experiment is shown in the code description.

\item {\bf Open access to data and code}
    \item[] Question: Does the paper provide open access to the data and code, with sufficient instructions to faithfully reproduce the main experimental results, as described in supplemental material?
    \item[] Answer: \answerYes{} 
    \item[] Justification: We provide detailed model parameter information in Appendix \ref{E} and attach reproducible code. And the dataset used in the experiment is shown in the code description.

\item {\bf Experimental Setting/Details}
    \item[] Question: Does the paper specify all the training and test details (e.g., data splits, hyperparameters, how they were chosen, type of optimizer, etc.) necessary to understand the results?
    \item[] Answer: \answerYes{} 
    \item[] Justification: We provide detailed model parameter information in Appendix \ref{E} and attach reproducible code. And the dataset used in the experiment is shown in the code description.

\item {\bf Experiment Statistical Significance}
    \item[] Question: Does the paper report error bars suitably and correctly defined or other appropriate information about the statistical significance of the experiments?
    \item[] Answer: \answerYes{} 
    \item[] Justification: As we demonstrated in Section \ref{section4}, we repeated the experiment ten times and reported the mean value and standard deviation of the results. In addition, instructions were provided on how to reproduce the results. 

\item {\bf Experiments Compute Resources}
    \item[] Question: For each experiment, does the paper provide sufficient information on the computer resources (type of compute workers, memory, time of execution) needed to reproduce the experiments?
    \item[] Answer: \answerYes{} 
    \item[] Justification: We introduced the comparison of computational efficiency in Section \ref{section4.2}(c), and the results are shown in Table \ref{table3}.
    
\item {\bf Code Of Ethics}
    \item[] Question: Does the research conducted in the paper conform, in every respect, with the NeurIPS Code of Ethics \url{https://neurips.cc/public/EthicsGuidelines}?
    \item[] Answer: \answerYes{} 
    \item[] Justification: Yes, our research complies with NeurIPS ethical standards.

\item {\bf Broader Impacts}
    \item[] Question: Does the paper discuss both potential positive societal impacts and negative societal impacts of the work performed?
    \item[] Answer: \answerNA{} 
    \item[] Justification: Our work belongs to basic research and has no significant social impact. 
    
\item {\bf Safeguards}
    \item[] Question: Does the paper describe safeguards that have been put in place for responsible release of data or models that have a high risk for misuse (e.g., pretrained language models, image generators, or scraped datasets)?
    \item[] Answer: \answerNA{} 
    \item[] Justification: the paper poses no such risks.

\item {\bf Licenses for existing assets}
    \item[] Question: Are the creators or original owners of assets (e.g., code, data, models), used in the paper, properly credited and are the license and terms of use explicitly mentioned and properly respected?
    \item[] Answer: \answerYes{} 
    \item[] Justification: All assets used in this paper have clear references and are publicly available.

\item {\bf New Assets}
    \item[] Question: Are new assets introduced in the paper well documented and is the documentation provided alongside the assets? 
    \item[] Answer: \answerYes{} 
    \item[] Justification: The code we release and the dataset we use are well documented in the code repository documentation.

\item {\bf Crowdsourcing and Research with Human Subjects}
    \item[] Question: For crowdsourcing experiments and research with human subjects, does the paper include the full text of instructions given to participants and screenshots, if applicable, as well as details about compensation (if any)? 
    \item[] Answer: \answerNA{} 
    \item[] Justification: the paper does not involve crowdsourcing nor research with human subjects.

\item {\bf Institutional Review Board (IRB) Approvals or Equivalent for Research with Human Subjects}
    \item[] Question: Does the paper describe potential risks incurred by study participants, whether such risks were disclosed to the subjects, and whether Institutional Review Board (IRB) approvals (or an equivalent approval/review based on the requirements of your country or institution) were obtained?
    \item[] Answer: \answerNA{} 
    \item[] Justification: the paper does not involve crowdsourcing nor research with human subjects.

\end{enumerate}

\end{document}